\documentclass{article} % For LaTeX2e

\usepackage{iclr2025_conference,times}
\usepackage{booktabs}
\usepackage{multirow}
\usepackage{algorithm}
\usepackage{graphicx}
\usepackage{array}
\usepackage{amsmath,amsfonts,amsthm}
\usepackage{mathtools}
\usepackage{lipsum}
\usepackage{dsfont}
\usepackage{caption}

\usepackage[noend]{algpseudocode}
\usepackage{amsmath,amsfonts,bm}

\usepackage{accents}
\newcommand{\ubar}[1]{\underaccent{\bar}{#1}}

\newcommand\normx[1]{\Vert#1\Vert}

\newtheorem{theorem}{Theorem}
\newtheorem{lemma}{Lemma}

\newtheorem{problem}{Problem}

\DeclareMathAlphabet{\mathcal}{OMS}{cmsy}{m}{n}
\SetMathAlphabet{\mathcal}{bold}{OMS}{cmsy}{b}{n}
\usepackage{dsfont}

\def\eqref#1{equation~\ref{#1}}
\def\1{\bm{1}}

\def\vg{{\bm{g}}}

\def\vq{{\bm{q}}}
\def\vdelta{{\bm{\delta}}}
\def\mDelta{{\bm{\Delta}}}

\def\vx{{\bm{x}}}

\def\mA{{\bm{A}}}

\def\mD{{\bm{D}}}

\def\mG{{\bm{G}}}

\def\mQ{{\bm{Q}}}

\def\mX{{\bm{X}}}

\def\mZ{{\bm{Z}}}

\DeclareMathAlphabet{\mathsfit}{\encodingdefault}{\sfdefault}{m}{sl}
\SetMathAlphabet{\mathsfit}{bold}{\encodingdefault}{\sfdefault}{bx}{n}

\DeclareMathOperator*{\argmax}{arg\,max}

\usepackage{hyperref}
\usepackage{url}

\definecolor{cadmiumgreen}{rgb}{0.0, 0.42, 0.24}
\definecolor{burgundy}{rgb}{0.5, 0.0, 0.13}

\title{\name{}: Lightweight Non-Parametric \\ Fine-Tuning of Embeddings for Retrieval}

\author{Sepanta Zeighami\\
UC Berkeley \\
\texttt{zeighami@berkeley.edu} \\
\And
Zac Wellmer \\
\texttt{zac@1984.ai}
\And
Aditya Parameswaran \\
UC Berkeley \\
\texttt{adityagp@berkeley.edu} \\
}

\newcommand{\name}{NUDGE}

\iclrfinalcopy % Uncomment for camera-ready version, but NOT for submission.
\begin{document}

\newcommand{\QG}[0]{{\mathcal{G}_{i, Y_i^V}- \mathcal{G}_{i, j}}}

\maketitle

\begin{abstract}
$k$-Nearest Neighbor search on dense vector embeddings ($k$-NN retrieval) from pre-trained embedding models is the predominant retrieval method for text and images, as well as Retrieval-Augmented Generation (RAG) pipelines. In practice, application developers often fine-tune the embeddings to improve their accuracy on the dataset and query workload in hand. Existing approaches either fine-tune the pre-trained model itself or, more efficiently, but at the cost of accuracy, train adaptor models to transform the output of the pre-trained model. We present \name{}, a family of novel \textit{non-parametric} embedding fine-tuning approaches that 
are significantly more accurate and efficient than both sets of existing approaches. %, while, respectively, running. %,  and across different choices of the pre-trained model. 
\name{} directly modifies the embeddings of data records to maximize the accuracy of $k$-NN retrieval. We present a thorough theoretical and experimental study of \name{}'s non-parametric approach. We show that even though the underlying problem is NP-Hard,
constrained variations can be solved efficiently. These constraints
additionally ensure that the changes to the embeddings are modest, avoiding large distortions to the semantics learned during pre-training. In experiments across five pre-trained models and nine standard text and image retrieval datasets, \textit{\name{} runs in minutes and often improves NDCG@10 by more than 10\% over existing fine-tuning methods. On average, \name{} provides 3.3$\times$ and 4.3$\times$ higher increase in accuracy and runs 200$\times$ and 3$\times$ faster, respectively, over fine-tuning the pre-trained model and training adaptors}. \footnote{Code available at \url{https://github.com/szeighami/nudge}}.
\end{abstract}

\if 0
\name{} is non-parametric, in that it modifies the embeddings themselves, rather 

However, fine-tuning large models computationally expensive and require access to model parameters, while the alternative approach of training adaptors requires less computational resources but also provides less accuracy improvements. In this paper, we present \name{} a family of non-parametric embedding fine-tuning approaches that improve accuracy more than fine-tuning the pre-trained model, and are more efficient than training adaptors. \name{} changes the data embedding, non-parametrically and within a constrained region, to maximize the similarity between data and query embeddings. Experimental results on standard text and image retrieval benchmarks show up to 24.4\% and on average 12.4\% accuracy boost from using \name{}, which also runs in minutes and does not require access to model parameters.
\fi

\section{Introduction}
$k$-Nearest Neighbor search on dense vector embeddings ($k$-NN retrieval) from pre-trained embedding models is the de-facto standard for text and image retrieval, as well as in Retrieval-Augmented Generation (RAG) pipelines. \citep{lewis2020retrieval, li2023blip, gao2023retrieval, patil2023gorilla, du2022amazon, liu2021pre}. Given  $n$ data records (e.g., text chunks or images), $k$-NN retrieval embeds them using a pre-trained model as $d$-dimensional vectors in $\mathcal{R}^d$. To answer a query, it similarly embeds the query in $\mathcal{R}^d$ and retrieves the top-$k$ data records whose embeddings have the highest cosine similarity (or inner product) with the query embedding. By simply performing a top-$k$ look-up (often through vector databases), the simplicity and efficiency of $k$-NN retrieval has made it increasingly popular and often preferred to other retrieval paradigms, e.g., late interaction \citep{khattab2020colbert, santhanam2021colbertv2} or generative retrieval \citep{tay2022transformer, wang2022neural}. However, the out-of-the-box pre-trained model is often not sufficiently accurate on the dataset or queries in hand, and typically fine-tuning is used to improve the accuracy. %This paper investigates how to efficiently fine-tune embeddings to boost the accuracy of $k$-NN retrieval. We consider a supervised setting with access to a training set of queries and their corresponding ground-truth data records that answer the queries. 

%MAYBE MENTION SOMEWHERE THAT THIS NEEDS Queries

There are two standard approaches to fine-tuning embeddings: fine-tuning the pre-trained models directly (referred to henceforth as \textit{PTFT}) or training \textit{adaptor models} on top of the pre-trained models \citep{zhao2024dense, zhou2024lima, huggingfaceTrainingFinetuning, trychromaEmbeddingAdapters, llamaindexFineTuningLlamaIndex}. PTFT can be more accurate but comes with practical challenges: it (1) requires access to the model parameters and must rely on third-party APIs for closed-source models \footnote{E.g., OpenAI currently does not provide an interface for fine-tuning embedding model \citep{openaifinetuning}.}, (2) is computationally expensive, and (3) incurs further hosting and maintenance costs for deployment of the fine-tuned model. An alternative is to learn an {\em Adaptor}, $\hat{g}_\theta$, a transformation of the output of the (frozen) pre-trained model, $\hat{f}$, %that is, to learn a suitable $\hat{g}_\theta$ 
so that the function $\hat{g}_\theta\circ\hat{f}$ generates accurate data and/or query embeddings. Training $\hat{g}_\theta$ can be done \textit{model-agnostically}, that is, with \textit{only black-box} access to the pre-trained model, addressing (1). Moreover, Adaptors are typically small models, such as linear models \citep{llamaindexFineTuningLlamaIndex}, addressing (2), and lowering the associated costs in (3). Nonetheless, experimental results show, at best, modest accuracy gains from using Adaptors. {\em We, therefore, lack a fine-tuning approach for $k$-NN retrieval that is simultaneously effective, efficient, and easy-to-use.} % compared with fine-tuning the pre-trained model.  

\begin{figure}
%\begin{minipage}{\textwidth}
\hspace{-1cm}
\begin{minipage}{0.55\textwidth}
    \centering
    \includegraphics[width=1\linewidth]{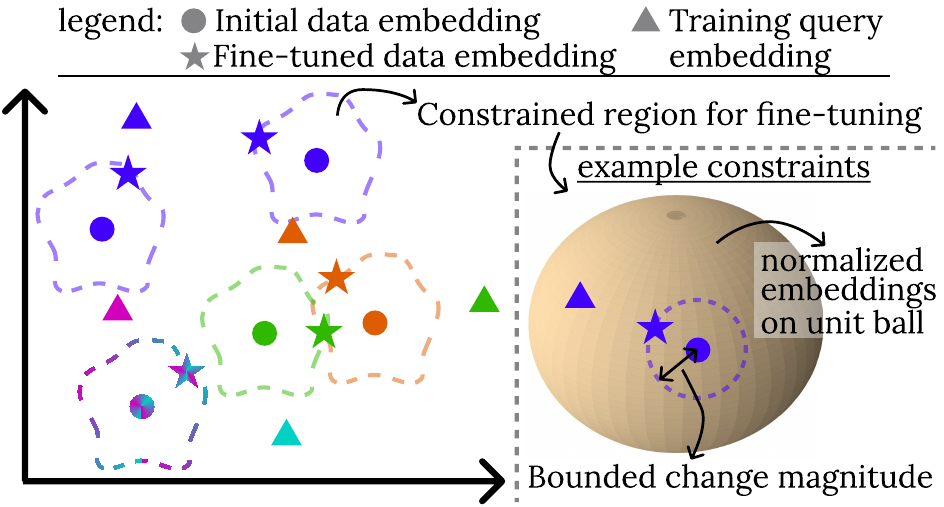}
    \captionof{figure}{\name{}s change embeddings within a constrained region to maximize similarity with training queries. Data embeddings are colored based on queries for which they are the ground-truth answers.}
    \label{fig:nopet_overview}
\end{minipage}    
%\begin{table}[t]
%\hspace{-1.5cm}
\hspace{0.1cm}
\begin{minipage}{0.6\textwidth}
\centering
        \begin{tabular}{>{\centering\arraybackslash}p{2.5cm} | c | c | >{\centering\arraybackslash}p{1.5cm}}
        %\toprule
        &\multirow{2}{*}{\textbf{PTFT}}&\multirow{2}{*}{\textbf{Adaptors}}&\textbf{\name{} (ours)}
   \\\hline%\midrule
\textbf{Needs Model Parameters}    & \multirow{2}{*}{Yes} & \multirow{2}{*}{No} & \multirow{2}{*}{\textbf{No}}\\\hline
%\textbf{Resource Requirement}  & \multirow{2}{*}{High}& \multirow{2}{*}{Low}& \multirow{2}{*}{\textbf{Low}}\\\hline
\textbf{Fine-Tuning Time (mins.)}  & \multirow{2}{*}{447}& \multirow{2}{*}{7.99}& \multirow{2}{*}{\textbf{2.18}}\\\hline
\textbf{Accuracy Boost (avg / max\%)}  & \multirow{2}{*}{3.8 / 15.6}& \multirow{2}{*}{2.9 / 12.4} & \multirow{2}{*}{\textbf{12.4 / 24.8}}%\\\bottomrule
\end{tabular}
    \captionof{table}{Comparison of fine-tuning methods on standard text datasets (PTFT refers to fine-tuning the pre-trained model). See Sec.~\ref{sec:exp} for experimental setting.}
    \label{tab:res_overview}
    \end{minipage}
%\end{table}
%\end{minipage}
\end{figure}

We present \name{}, a family of approaches to \textit{\underline{n}on-parametrically fine-t\underline{u}ne embe\underline{d}din\underline{g}s \underline{e}fficiently.} %. \name{} presents family of \textit{non-parametric} approaches to embedding fine-tuning 
\name{} methods (or \textit{\name{}s}) are surprisingly effective, model-agnostic, and incur no additional deployment cost. \name{}s take a novel non-parametric view of embedding fine-tuning: they view the embeddings themselves as parameters of the $k$-NN retrieval algorithm and directly modify the embeddings of data records to maximize the accuracy of $k$-NN retrieval. Although %Nonetheless, doing so efficiently while generalizing beyond the training set is nontrivial. 
we show that the underlying optimization problem is NP-Hard in general, and can lead to overfitting if data embeddings are allowed to change arbitrarily, \name{}s efficiently solve constrained variations of this problem formulated to avoid overfitting. As shown in Fig.~\ref{fig:nopet_overview}, \name{}s change each data embedding to maximize the similarity between the data embedding and the training queries to which the data record is a correct answer, while constraining how and by how much the embedding can change. Intuitively, the constraints allow for enough modifications to the embeddings to improve accuracy on the dataset in hand while avoiding large distortion that would offset the semantics learned during pre-training. Fig.~\ref{fig:nopet_overview} shows an example of the constrained region used, where new embeddings are constrained to be normalized (i.e., to fall on the unit ball), and the magnitude of changes to the embeddings to be bounded. \name{}s solve the constrained optimization problems in closed form, presenting simple and effective update formulae for embedding fine-tuning. 
%\name{}s change each data embedding to within a constrained region to 
%\name{}s furthermore show how to set the bound on embedding change magnitude optimally to maximize validation accuracy, presenting closed-form solutions that are both accurate and extremely efficient.

\name{}'s non-parametric formulation %improves both accuracy and efficiency. It 
contrasts with the parametric approaches adopted by PTFT and Adaptors that increase the similarity between query and answer pairs through contrastive-type losses \citep{zhao2024dense}. %, \name{}s change a data embedding by moving it to increase its similarity to queries to which it's a correct answer. 
Through constrained non-parametric optimization, \name{}s instead make \textit{bounded local} changes to each individual data embedding. % \textit{independently} of other embeddings. %This avoids large changes to the embeddings for data records that are underrepresented in the training set. 
On the other hand, PTFT and Adaptors make \textit{potentially large global} changes to the embedding function that modify the embeddings of \textit{all possible} data and queries. However, learning modifications that generalize across all data and queries is difficult from the typically \textit{small} training sets available during fine-tuning. Instead, models end up overfitting to the training set. Besides accuracy gains, the non-parametric formulation allows for efficient closed-form solutions to the optimization problem. \name{}s use computation equivalent (up to a log factor) to a \textit{single} training and validation iteration of Adaptors. %for fine-tuning.% embeddings. %Thus, the accuracy gains from \name{}s are paired with significant efficiency boosts.% compared with parametric approaches. 

We present a thorough experimental evaluation, showing that \name{}s significantly outperform Adaptors and PTFT on standard text retrieval datasets from BEIR \citep{thakur2021beir} and KILT \citep{petroni-etal-2021-kilt}
and image retrieval datasets COCO~\citep{lin2014microsoft} and Flickr~\citep{young2014image}. \name{}s improve common metrics such as Recall@$10$ and NDCG@$10$ by \textit{up to 16.0\% more than both PTFT and Adaptors and 24.4\% over no fine-tuning. Across 9 different datasets and 5 different embedding models \name{}s consistently outperform PTFT and Adaptors, often providing 10\% more increase in NDGC@10.} % by 11.7\% over no fine-tuning, compred with  2.66\% of Adaptors and 3.6\% (on a smaller subset of datasets) of PTFT.} 
\name{}s avoid performance degradation on out-of-distribution queries that affect parametric approaches, ensuring that the significant accuracy gains for in-distribution queries do not come at the expense of out-of-distribution queries. 
%up to 13\% more compared with existing fine-tuning methods, and provide up to 16\% accuracy gains over using non-fine-tuned embeddings}. 
Moreover, \name{}s are model-agnostic and can be used with any (potentially closed-source) embedding model. Overall, \name{}s are trained in minutes on datasets with millions of records, and the fine-tuned embeddings can be used directly without any additional model inference cost, providing significant accuracy boosts almost ``for free''. 

\textbf{Contributions and Overview}. To summarize, our contributions are as follows. 
\begin{itemize}
    \item We formalize the notion of non-parametric embedding fine-tuning and show that the underlying unconstrained non-parametric optimization problem is NP-Hard 
    (Sec.~\ref{sec:method:problem}).
    \item We present \name{}-M and \name{}-N, two methods that optimally solve constrained variations of the problem (Sec.~\ref{sec:method:solution}).
    \name{}-M updates data embeddings to maximize the similarity between the embeddings of queries and their ground-truth data records subject to a bound on the magnitude of change to the embeddings.
    \name{}-N adopts a similar approach, but additionally constrains the embeddings to be normalized. 
    \item We show \name{} variants consistently outperform parametric fine-tuning methods with thorough experiments on 5 embedding models and 9 standard retrieval datasets  (Sec.~\ref{sec:exp}).
\end{itemize}

\section{Preliminaries}\label{sec:prelim}
\textbf{Notation}. We use bar $\bar{}$ on top of letters to denote raw data/queries that are not embedded. We use boldface capital letters to denote matrices (mostly used to represent embeddings), e.g., $\mX\in\mathcal{R}^{r\times s}$ is an $r\times s$ matrix, and use $\mX_i$ to refer to the $i$-th row of the matrix, which is a vector, $\mX_i\in\mathcal{R}^s$. $\normx{\vx}$ refers to the L2 norm of a vector $\vx$, and $\normx{\mX}$ refers to the vector of row-wise norms,  $(\normx{\mX_1}, ..., \normx{\mX_r})$. We use $[s]$ to refer to the set $\{1, 2, ..., s\}$. 

%\subsection{Problem Setting}
\textbf{$k$-Nearest Neighbor Retrieval}. Given a dataset, $\bar{D}$ of $n$ records (e.g., text chunks or images), $k$-NN retrieval first embeds the data records using an embedding model, $\mathcal{E}_D$, to generate an embedding matrix $\mD\in \mathcal{R}^{n\times d}$ where $d$ is the embedding dimensionality. The $i$-th row, $\mD_i$, of $\mD$ is the embedding of the $i$-th record, $\bar{D}_i$, that is, $\mD_i=\mathcal{E}_D(\bar{D}_i)$. To answer a query $\bar{q}$, $k$-NN retrieval first embeds the query using an embedding model, $\mathcal{E}_Q$ (often $\mathcal{E}_Q$ and $\mathcal{E}_D$ are the same, but can be different, especially in a multi-modal setting)  to obtain $\vq=\mathcal{E}_Q(\bar{q})$. $k$-NN retrieval then returns the top-$k$ records in $\bar{D}$ whose embeddings have the highest similarity to $\vq$, measured by either the inner product or cosine similarity, i.e., the $k$ records $\bar{D}_{{i}_1}, ..., \bar{D}_{{i}_k}$ that correspond to the $k$ highest values in the set $\{\mD_1\cdot\vq, ..., \mD_n\cdot\vq\}$ when using the inner product similarity metric. By default, we use the inner product, which can be applied to normalized embeddings to obtain cosine similarity. %Pre-trained models are often used out of the box to obtain the embeddings, and the embeddings are often stored in vector databases to speed-up $k$-nearest neighbour search on the fixed database.

%We are given a dataset $D$ of $n$ records (e.g., text chunks or images), and a data embedding model $\mathcal{E}_D$ that given a record $r\in D$ outputs $d$-dimensional embedding of $r$. We are also given a query embedding model $\mathcal{E}_Q$ (often the same as $\mathcal{E}_D$, but can be different in a multi-modal setting), that given a query $q$ outputs a $d$-dimensional embedding of $q$. A $k$-nearest neighbor retrieval method, for a query $q$ returns $k$ records in $D$ with maximum similarity score, where, $s$, the similarity score of $q$ and $r\in D$ is either their dot product or cosine similarity, which is the dot product of $\frac{q_e}{\normx{q_e}}$ and $\frac{r_e}{\normx{r_e}}$. The goal is to ensure the true answer to $q$ is in the set of $k$ documents retrieved based on the retrieval method XXXXX.  XXX

\textbf{Ground-Truth Answers}. Queries can require retrieving multiple data records, and each data record can have a different degree of relevance to the query.  For simplicity, here, we present our results in the setting where a query, $\bar{q}$, requires retrieving a single ground-truth data record. We refer to the ground-truth data record for a query, $\bar{q}$, as the \textit{ground-truth answer} to the query and often refer to the data record with its index $y$, $y\in[n]$. Extensions to multiple ground-truth data records (each with potentially different degrees of relevance to the query) is straightforward and presented in Appx.~\ref{appx:multi_label}. % extends our results, in a straightforward manner, to queries with multiple ground-truth answers, where each answer can have a potentially different degree of relevance to the query. % where each answer can potentially have a different degree of relevance to the query. %, as discussed . %defer the discussion on queries 

\textbf{Fine-Tuning}. Fine-tuning aims to improve retrieval accuracy for the dataset $\bar{D}$ through optimizing embeddings. We let $\mD^*\in\mathcal{R}^{n\times d}$ denote the fine-tuned data embeddings obtained after fine-tuning, where $\mD^*_i\in\mathcal{R}^d$ is the fine-tuned embedding for the $i$-th data record. We consider a supervised setting where a query set with corresponding ground-truth answers is available, consisting of the query set, $\bar{Q}$, and a set, $Y$, of ground-truth answers, where $Y_j$ is the index of the ground-truth answer for the $j$-th query, $\bar{q}_j$. We split this set into two, a training set $\bar{Q}^T, Y^T$ and a validation set $\bar{Q}^V, Y^V$, with $n_T$ and $n_V$ queries, respectively. Let $\mQ^T\in\mathcal{R}^{n_T\times d}$ and $\mQ^V\in\mathcal{R}^{n_V\times d}$ be matrices containing embeddings for training and validation queries. %For simplicity, we present most of our results assuming there is only one ground-truth answer or data record for any query, $q_j$, which we call the  \textit{correct answer} to $q_j$. We discuss extensions to multiple correct answers with relevance scores being available in Appx.~\ref{appx:multi_label}. 
The training set can be collected over time from user interactions with the system, by collecting labels, or by generating synthetic training data using LLMs (e.g., \cite{llamaindexDatasetGeneration, meng2022generating}).

\section{Non-Parametric Embedding Fine-Tuning}
Our \textit{\name{}} approach views embeddings as parameters of the $k$-NN retrieval algorithm, optimizing the embeddings directly to improve retrieval accuracy. %This is done by finding a suitable vector, $\vdelta_i\in\mathcal{R}^d$, to move the embedding of the $i$-th data record, $\mD_i$, to improve the $k$-NN retrieval accuracy. 
In Sec.~\ref{sec:method:problem}, we formalize the notion of non-parametric embedding fine-tuning by stating two optimization problems, one directly maximizing a retrieval accuracy metric and one maximizing similarity between queries and their ground-truth answers. Maximizing accuracy is the final goal, but similarity is a simpler surrogate to optimize in practice. Nonetheless, we show that the former is NP-hard and the latter is unbounded (i.e., the objective can be improved indefinitely). Moreover, since both optimization problems allow embeddings to arbitrarily change (i.e., are unconstrained), directly solving either problem can lead to overfitting. In Sec.~\ref{sec:method:solution}, we present \name{}, a family of approaches that solve a combination of constrained variations of the two optimization problems to address generalization and efficiency challenges. %Sec.~\ref{sec:method:practice} discusses various practical considerations and variations of \name{}.

\subsection{Unconstrained Non-Parametric Embedding Fine-Tuning Problems}\label{sec:method:problem}
%We present two problem statements, one that maximizes retrieval \textit{accuracy} and one that maximizes \textit{similarity} between correct query/answer pairs. %\name{} combines the two optimization problems: maximizing accuracy on a validation set while maximizing similarity on a training set. 
%Both problem statements are \textit{non-parametric}, that is, the embeddings are not parameterized (as they are in adaptor-based 
%approaches) but instead, optimized by finding a suitable vector that modifies them. 
Let $\mDelta\in\mathcal{R}^{n\times d}$ be the modification to be learned to the embeddings, so that its $i$-th row, $\mDelta_i$, is the modification to $\mD_i$. That is, after fine-tuning, the final embedding is $\mD^*=\mD+\mDelta$. We use $\mQ$ and $Y$ to refer to a generic query embedding matrix with corresponding ground-truth answers containing $N$ queries. $\mQ$ and $Y$ can respectively be either $\mQ^T$ and $Y^T$ or $\mQ^V$ and $Y^V$.
 
\textbf{MaxA-EFT}. \textit{Maximum Accuracy Embedding Fine-Tuning Problem, MaxA-EFT}, is the problem of finding $\mDelta$ to fine-tune data embeddings that maximizes the number of queries in $\mQ$ answered correctly, % \textit{accuracy with respect to the query set }, 
formalized as follows. For a query embedding matrix $\mQ$ with ground-truth answers $Y$, let $\mathcal{I}_i(\mDelta)$, for $i\in [N]$, be the indicator variable denoting if the $i$-th query is \textit{answered correctly after fine-tuning with $\mDelta$}. Formally, $\mathcal{I}_i(\mDelta)=1$ if the following holds, and zero otherwise%the $i$-th training query is answered correctly after modifying the data embeddings by $\mDelta$, or formally, when the inequality 
\begin{align}\label{eq:I_iT}
    \mQ_i\cdot(\mD_{Y_i}+\mDelta_{Y_i}) > \mQ_i\cdot(\mD_{j}+\mDelta_{j}),\quad\forall j\in[n]\setminus Y_i.
\end{align}
    %holds. We say that  when $\mathcal{I}_i(\mDelta)=1$. %We define accuracy as the total number of queries in the query set answered \textit{correctly}, formalized as follows. 
\if 0
We say that the $i$-th query, $i\in[N]$, is \textit{answered correctly after fine-tuning with $\mDelta$} if $\forall j\in[n]\setminus Y_i$
\begin{align}\label{eq:single_q_accuracy}
    \mQ_i\cdot(\mD_{Y_i}+\mDelta_{Y_i}) > \mQ_i\cdot(\mD_{j}+\mDelta_{j}).
\end{align}
%Using Eq.~\ref{eq:single_q_accuracy}, we formally state 
MaxA-EFT is the problem of finding $\mDelta$ to maximize the number of queries in $\mQ, Y$ \textit{answered correctly} after fine-tuning:% as a maximum constraint satisfaction problem as follows.
\fi
\begin{problem}[MaxA-EFT]
MaxA-EFT is the problem of finding $\mDelta$ to maximize the number of queries answered correctly after fine-tuning with $\mDelta$, i.e., ${\arg\max}_{\mDelta\in\mathcal{R}^{n\times d}}\sum_{i\in[N]}\mathcal{I}_{i}(\mDelta).$
%$$\underset{\mDelta\in\mathcal{R}^{n\times d}}{\arg\max}\sum_{i\in[N]}\mathcal{I}_{i}(\mDelta).$$
    %is equivalent to finding $\mDelta$ that satisfies the maximum number of $\mathcal{I}_i(\mDelta)$ variables, so that MaxA-EFT is formally the optimization problem 
\end{problem}

\begin{theorem}\label{thm:np_hard}
    MaxA-EFT is NP-Hard.
\end{theorem}
Theorem~\ref{thm:np_hard} is proved by reduction from the Maximum Feasible Linear Subsystem problem as studied in ~\cite{amaldi1995complexity}, see Appx.~\ref{appx:np_hard_proof}. Apart from the NP-hardness, MaxA-EFT allows data embeddings to be arbitrarily changed by $\mDelta$. This can distort the semantics captured in $\mD$ by the pre-trained model, and lead to poor generalization to queries outside of $\mQ$. %overfitting when the training set is small. 
%For instance, in the special case of each data record being the correct answer to a single query, the optimal solution is to simply set the data embedding equal to the embedding of the query for which the data record is the correct answer. This change essentially discards any information in the data record not related to the observed queries. %Avoiding overfitting and improving computational efficiency are the central themes of our solution, \name{}.

\textbf{MaxS-EFT}. An alternative formulation is \textit{Maximum Similarity Embedding Fine-Tuning Problem,}
\begin{align}\label{eq:max_sim}
    \underset{\mDelta\in\mathcal{R}^{n\times d}}{\arg\max} \sum_{i\in[N]}\mQ_i\cdot (\mD_{Y_i}+\mDelta_{Y_i}),
\end{align}
referred to as \textit{MaxS-EFT}. %\textit{MaxS-EFT} fine-tunes embeddings to maximize similarity between queries and their correct answer, instead of maximizing accuracy. x
Here, we change data embeddings to maximize the similarity between queries and ground-truth answers,  
%fine-tune embeddings to maximize similarity between queries and their correct answer. 
a standard optimization objective (e.g., \cite{henderson2017efficient}, \cite{sbertLossesx2014cos}). However, the non-parametric formulation makes Eq.~\ref{eq:max_sim} an unconstrained optimization problem with a linear objective, so the problem is unbounded and has no optimal solution. Moreover, setting $\mDelta_{Y_i}$ so that $\mQ_i\cdot \mDelta_{Y_i}>0$ and increasing the magnitude of $\mDelta_{Y_i}$ arbitrarily improves the objective, yielding trivial solutions with poor generalization to unseen queries. %As such, directly optimizing the objective Eq.~\ref{eq:loss} can lead to overfitting.

%Thus, this problem, similar to MaxA-EFT, cannot be solved to optimality. Moreover, similar to MaxA-EFT, allowing the embeddings to be drastically changed by $\mDelta$ can lead to overfitting. For instance, regularization by adding the constraint $\normx{\mD_i+\mDelta_i}_2=1$ (to ensure fine-tuned embeddings are normalized), makes the problem bounded. However, in the special case where each data record is the correct answer to a single query, the optimal solution is to set $\mDelta_i$ so that $\mD_i+\mDelta_i$ is the normalized embedding of the query whose correct answer is $\mD_i$, overfitting to the training set. 

\subsection{\name{} Approaches}\label{sec:method:solution}
%To summarize, Sec.~\ref{sec:method:problem} formalized non-parametric fine-tuning, and showed two associated challenges: (1) avoiding overfitting and (2) ensuring efficiency. 
Because of the potential for overfitting and the computational challenges due to NP-hardness, we do not solve either MaxA-EFT or MaxS-EFT directly. Instead, we introduce \name{}, a family of approaches that solve constrained variations of MaxA-EFT and MaxS-EFT, designed to avoid overfitting, while being efficient. We discuss two main approaches, \name{}-M and \name{}-N, in Secs.~\ref{sec:nopet_m} and \ref{sec:nopet_n} and present other practical extensions in Appx.~\ref{sec:method:practice}.  %Sec.~\ref{sec:nopet_m}-\ref{sec:method:practice} present different  \name{} variations, each considering a different set of constraints. 

%In Sec.~\ref{sec:nopet_m}, we first present \  textit{\name{}-M}, the \textit{bounded magnitude} version of \name{} that bounds $\normx{\mDelta_i}$'s to ensure fine-tuning does not change the data embeddings too much. We then present \textit{\name{}-N}, which additionally constrains the fine-tuned embeddings to be \textit{normalized}. Finally, Sec.~\ref{sec:method:practice} presents other potential variations.  

\subsubsection{\name{}-M: \name{} with Bounded Magnitude}\label{sec:nopet_m}
\name{}-M solves MaxS-EFT on the training set, but with the added constraint $\normx{\mDelta_i}\leq \gamma$, $\forall i\in [n]$, for a scalar $\gamma\geq 0$. $\gamma$ controls how much each embedding can change during fine-tuning. \name{}-M sets $\gamma$ by solving MaxA-EFT on the validation set. %Let $\mDelta^\gamma$ be an optimal solution to the constrained MaxS-EFT on the training set. \name{}-M then sets $\gamma$ so that $\mDelta^\gamma$ maximizes the MaxA-EFT objective on the validation set. 
Intuitively, this (1) changes data embeddings to maximize the similarity between embeddings and queries on the training set, (2) ensures that the magnitude of the changes to the embeddings is bounded to avoid overfitting, and (3) decides how much the embeddings are allowed to change by maximizing validation accuracy. \name{}-M provides a closed-form solution to do this. %XXXX \name{}-M is an algorithm that optimally solves BiMax-M. 
We provide an overview of the solution here and leave the details and formal proofs to Appx.~\ref{proof:them:nopetm}.

\textbf{Optimization Formulation}. Define \texttt{MaxS-M}($\gamma$), for a scalar $\gamma\geq0$, as the set of optimal solutions to the following constrained version of MaxS-EFT with \textit{bounded magnitude}:%, referred to as \textit{MaxS-M}:  
\begin{align*}%\label{eq:max_bi}
    \texttt{MaxS-M}(\gamma)=\argmax & \sum_{i\in[n_T]}\mQ^T_i\cdot (\mD_{Y_i^T}+\mDelta_{Y_i^T})\\
    \textnormal{s.t.}&\;\,\mDelta\in\mathcal{R}^{n\times d}, \normx{\mDelta_i}\leq \gamma\quad \forall i\in[n].
\end{align*}
A $\mDelta\in\texttt{MaxS-M}(\gamma)$ changes data embeddings by at most $\gamma$ while maximizing the similarity between training queries and their ground-truth answer. We set $\gamma$ to maximize the validation accuracy after fine-tuning with $\mDelta\in\texttt{MaxS-M}(\gamma)$: %stated as the following bi-level optimization problem:

%Define the \textit{, BiMax-MEFT} as:%following bi-level optimization problem. %, solving the constrained  MaxS-EFT on a training set while solving MaxA-EFT on a validation set, and provides a closed-form solution to this optimization problem. %, finding the optimal value for $\gamma$ without the need to perform any hyperparameter tuning. 

%\textbf{\name{}}. First, we formally define Constrained BiMax-EFT, a constrained bi-level optimization problem that combines MaxA-EFT and MaxS-EFT, defined as 
\begin{align*}%\label{eq:max_bi}
    \max &\sum_{i\in[n_V]}\mathcal{I}_{i}^V(\mDelta)\\    \textnormal{s.t.\quad}&\mDelta\in\texttt{MaxS-M}(\gamma), \gamma\geq 0.\\
\end{align*}
$\mathcal{I}_i^{V}(\mDelta)$ denotes $\mathcal{I}_i$ (see Eq.~\ref{eq:I_iT}) on the validation set $\mQ^V,Y^V$, so $\sum_{i\in[n_V]}\mathcal{I}_{i}^V(\mDelta)$ is the validation accuracy after fine-tuning with $\mDelta$. %We call this optimization problem . 
%\end{problem}.
This problem is referred to as \textit{Bi-level Maximization with bounded Magnitude, BiMax-M}. We denote the optimal solution to BiMax-M by $\mDelta^M$. %Recall that given $\mDelta^M$, fine-tuned embeddings are obtained as $\mD^*=\mD+\mDelta^M$, so that computing fine-tuned embeddings is straightforward after finding $\mDelta^M$. %We use $\mDelta^*$ to refer to the optimal solution to BiMax-M, which is a $\mDelta^\gamma$ for an optimal $\gamma$. 

\textbf{\name{}-M}. \name{}-M is an algorithm that optimally solves BiMax-M:
\begin{theorem}\label{thm:nopet}
    There exists an algorithm, referred to as \name{}-M, that optimally solves BiMax-M in $O(n_V(nd+ \log n_V)+n_Td)$. Specifically, \name{}-M sets $\mDelta^M$ as%\name{}-M sets $\mD^*=\mD+\mDelta^M$ with $\mDelta^{M}$ %from Eq.~\ref{eq:nopetm:delta_gamma}.%, for an optimally chosen $\gamma^*$. % from Eq.~\ref{eq:nopetm:gamma_star}.
    \begin{align}\label{eq:nopetm:delta_gamma}
\mDelta_i^{M}=\gamma^*\frac{\mG_i}{\normx{\mG_i}},\quad\text{where}\;\; \mG_i= \sum_{j\in[n_T]}\mathds{I}[i=Y_j^T]\mQ^T_j, \quad \forall i\in [n], 
\end{align}
and $\mathds{I}$ is the indicator function with a predicate argument, for an optimally chosen scalar $\gamma^*$.
\end{theorem}
Eq.~\ref{eq:nopetm:delta_gamma} presents the simple update rule used by \name{}-M to fine-tune data embeddings, using which the fine-tuned embeddings are computed as $\mD^*=\mD+\mDelta^M$. Observe that $\mG_i$ is the sum of query embeddings whose ground-truth answer is $\mD_i$, so that data embeddings are moved towards the queries for which they are the ground-truth answers. We also note that $O(n_Td)$ is the complexity of a single iteration over training data, and $O(n_Vnd)$ is the complexity of calculating validation accuracy once. Thus, ignoring the log term, the above time complexity is equal to a single training and validation iteration for parametric approaches (i.e., Adaptors or PTFT), and has smaller constant factors since it does not perform any model forward passes. 

The optimal $\gamma^*$ in Eq.~\ref{eq:nopetm:delta_gamma} is calculated by solving linear inequalities resulting from the definition of $\mathcal{I}_{i}^V$. We provide an overview here and leave the details to Appx.~\ref{proof:them:nopetm}. First, note that $\texttt{MaxS-M}(\gamma)=\gamma\frac{\mG}{\normx{\mG}}$, found using the KKT points of the optimization problem. Thus, BiMax-M reduces to finding a $\gamma$ that maximizes $\sum_{i\in[n_V]}\mathcal{I}_{i}^V(\gamma\frac{\mG}{\normx{\mG}})$.
%Eq.~\ref{eq:nopetm:delta_gamma} is the optimal solution to $\texttt{MaxS-M}(\gamma)$ found using the KKT points of the optimization problem, as shown in Appx.~\ref{proof:them:nopetm}. 
For each $i\in[n_V]$, substituting $\mDelta=\gamma\frac{\mG}{\normx{\mG}}$ into the definition of $\mathcal{I}_{i}^V(\mDelta)$ in Eq.~\ref{eq:I_iT}, we have $\mathcal{I}_{i}^V(\gamma\frac{\mG}{\normx{\mG}})=1$ if the following holds, and zero otherwise:%inequalities in Eq.~\ref{eq:I_iT} are satisfied and is zero otherwise. Thus, , we obtain a set of linear inequalities in $\gamma$:
\begin{align}\label{eq:I_iV}
    \mQ_i^V\cdot(\mD_{Y_i}+\gamma\frac{\mG_{Y_i^V}}{\normx{\mG_{Y_i^V}}}) > \mQ_i^V\cdot(\mD_{j}+\gamma\frac{\mG_{j}}{\normx{\mG_{j}}}),\quad\forall j\in[n]\setminus Y_i^V.
\end{align}
Eq.~\ref{eq:I_iV} is a set of inequalities in $\gamma$. Denote the solution to the inequalities by $I_i$, so that $\gamma\in I_i$ if and only if $\mathcal{I}_{i}^V(\gamma\frac{\mG}{\normx{\mG}})=1$. Since the inequalities are linear in $\gamma$, $I_i$ is an interval in $\mathcal{R}$. Consequently, 
\begin{align}\label{eq:find_gammastar}
    \gamma^*=\argmax_{\gamma} \sum_{i\in n_V}\mathds{I}[\gamma\in I_i],
\end{align} 
and the solution to Eq.~\ref{eq:find_gammastar} is a $\gamma$ that intersects the most number of intervals across all $I_i$, $i\in[n_V]$, which can be found by a single iteration over the intervals after sorting their start and end points. %We defer the details of this approach to Appx.~\ref{proof:them:nopetm}, where we show:

%that To find the optimal $\gamma$, observe that $\mDelta^\gamma$ is linear in $\gamma$, so that substituting $\mDelta^\gamma$ into definition of $\mathcal{I}^V_i$ in Eq.~\ref{eq:I_iT} provides a set of linear inequalities in $\gamma$. To maximize $\sum_{i\in[n_V]}\mathcal{I}_{i}^V(\mDelta^{\gamma})$, \name{}-M finds a $\gamma$ that maximizes the total number of such linear inequalities satisfied. 

%See Appx.~\ref{proof:them:nopetm} for details. 

%\textit{Proof sketch.}  \qed

%\name{}-M follows the above procedure, first computing $\mG$ which takes $O(n_Td)$ number of operations, then finding $I_i$ for all $i\in n_V$, which takes $O(n_Vnd)$, and finally using $I_i$ to calculate the optimal $\gamma$ which requires $O(n_V\log n_V)$ to sort the intervals and perform a single pass over them. 

\subsubsection{\name{}-N: \name{} with Normalized Embeddings}\label{sec:nopet_n}
\name{}-N additionally constrains the norm of the fine-tuned embedding. This constraint serves as an additional regularization, helping with out-of-distribution generalization (see Sec.\ref{sec:exp:dist_shift}). 

\textbf{Optimization Formulation}. Analogous to $\texttt{MaxS-M}(\gamma)$ but with an added constraint, we define $\texttt{MaxS-N}(\gamma)$ as the optimal solution to the following optimization problem:%, referred to as \textit{MaxS-N}:
\begin{align*}%\label{eq:max_bi}
    \texttt{MaxS-N}(\gamma)=\argmax & \sum_{i\in[n_T]}\mQ^T_i\cdot (\mD_{Y_i^T}+\mDelta_{Y_i^T})\\
    \textnormal{s.t.}&\;\,\mDelta\in\mathcal{R}^{n\times d}, \normx{\mDelta_i}^2\leq \gamma, \textcolor{cadmiumgreen}{\normx{\mD_i+\mDelta_i}= 1}\quad \forall i\in[n].
\end{align*}
To find a suitable $\gamma$, we solve the same optimization problem as BiMax-M, except that we replace $\texttt{MaxS-M}(\gamma)$ with $\texttt{MaxS-N}(\gamma)$. We call this problem \textit{Bi-level Maximization with Normalized embeddings, BiMax-N}, and refer to its optimal solution as $\mDelta^N$. %, with the final fine-tuned embeddings $\mD^*=\mD+\mDelta^N$

\textbf{\name{}-N}. \name{}-N is an algorithm that optimally solves BiMax-N:%. The optimal embedding update \textbf{\name{}-M}. \name{}-M is an algorithm that optimally solves BiMax-M:
\begin{theorem}\label{thm:nopet_n}
    There exists an algorithm, referred to as \name{}-N, that optimally solves BiMax-N in $O(n_V(nd+ \log n_V)+n_Td)$. Specifically, \name{}-N sets $\mDelta^N$ as%\name{}-M sets $\mD^*=\mD+\mDelta^M$ with $\mDelta^{M}$ %from Eq.~\ref{eq:nopetm:delta_gamma}.%, for an optimally chosen $\gamma^*$. % from Eq.~\ref{eq:nopetm:gamma_star}.
\begin{equation}\label{eq:nopetn_delta_gamma}
    \mDelta_i^N=\begin{cases*}
    \frac{\mG_i}{\normx{\mG_i}}-\mD_i,              & $\textnormal{if}\; \frac{\mG_i\cdot\mD_i}{\normx{\mG_i}}\geq 1-\frac{\gamma^*}{2}$\\
    \frac{\sqrt{\gamma^*(4-\gamma^*)}}{2}\mZ_i-\frac{\gamma^*}{2}\mD_i,              & \textnormal{otherwise,}
\end{cases*}
\end{equation}
where 
$$\mZ_i=\frac{\mG_i-(\mD_i\cdot\mG_i)\mD_i}{\normx{\mG_i-(\mD_i\cdot\mG_i)\mD_i}},$$
and $\mG_i$ is as defined in Eq.~\ref{eq:nopetm:delta_gamma}, for an optimally chosen scalar $\gamma^*$.
    \end{theorem}
%performed by \name{}-N is %, an algorithm that optimally solves BiMax-N. 
%To find $\Delta^\gamma=\texttt{MaxS-N}(\gamma)$, observe that using KKT conditions, as shown in Appx.~\ref{proof:them:nopetn}, yields, for any $i\in[n]$,
Eq.~\ref{eq:nopetn_delta_gamma} is the update rule used by \name{}-N to fine-tune data embeddings, using which the fine-tuned embeddings are obtained as $\mD^*=\mD+\mDelta^{N}$. This moves the data embedding on the unit ball (to satisfy $\normx{\mD_i+\mDelta_i}= 1$) between $\mD_i$ and $\frac{\mG_i}{\normx{\mG_i}}$, where $\frac{\mG_i}{\normx{\mG_i}}$ is the normalized sum of embeddings of queries whose ground-truth answer is $\mD_i$. $\gamma^*$ determines how much to move the embedding, and $\mZ_i$ determines the direction. %$\mG_i$ is the gradient of the objective of $\texttt{MaxS-N}$, and that 
$\mZ_i$ is the normalized projection of $\mG_i$ onto the tangent plane of the unit ball at $\mD_i$, so that moving in the direction of $\mZ_i$ maximally increases $\texttt{MaxS-N}$ objective. %, since $\mG_i$ is the gradient of the objective with respect to $\mDelta_i$. 

Finding the optimal $\gamma^*$ is similar to \name{}-M, where we first use KKT to solve $\texttt{MaxS-N}(\gamma)$ and substitute the resulting $\mDelta$ into the definition of $\mathcal{I}^V_i$ in Eq.~\ref{eq:I_iT}. The optimal $\gamma^*$ is then found as the value that satisfies the maximum number of the resulting inequalities. The resulting inequalities are, in this case, quadratic due to $\sqrt{\gamma(4-\gamma)}$ in the solution to $\texttt{MaxS-N}(\gamma)$, but can still be solved in closed-form. %In Appx.~\ref{proof:them:nopetn}, we show:
%Although updating embeddings by $\mDelta^N$ as in Eq.~\ref{eq:nopetn_delta_gamma} is simple, the process of finding the optimal $\gamma^*$ value can be tedious as it requires considering various special cases when solving quadratic equations. 
Nevertheless, from a practical perspective, solving the quadratic equations is tedious as it requires considering various special cases. We observed that performing a grid search to find $\gamma^*$ that maximizes validation accuracy finds good enough solutions and is almost as efficient. Thus, in our experiments, we use this practical implementation. %As our experiments in Sec.~\ref{sec:exp:ablation} show, \name{}-N is not sensitive to the value of $\gamma^*$, and similar values of $\gamma^*$ work well even across datasets.  

\if 
IS THERE A REGULAIZATION INTERPRETATION OF THIS?
A central problem for both MaxA-EFT and MaxS-EFT is overfitting. \name{} uses a form of cross-validation to address this (optionally coupled with normalization, see Sec.~\ref{sec:method:practice}). 
In doing so, \name{} attempts to solve neither MaxA-EFT nor MaxS-EFT to optimality, but solves a combination of the two problems. %Instead, it solves MaxS-EFT by gradient descent and selects but model tha. 
\textit{Conceptually}, \name{} can be seen as performing gradient descent on loss $\mathcal{L}$ (i.e., maximizing MaxS-EFT objective on the training set), but with the number of training steps and learning rate determined to maximize MaxA-EFT objective on a validation set. 
%This is,  by iteratively performing gradient descent on a training set on loss $\mathcal{L}$, while checking validation accuracy  after every update, and selecting the model at the epoch and learning rate with the highest validation accuracy. %Indeed a simple implementation of this as described above is possible and already performs well (see Sec.XX). 
However, \name{} sees the above as a bi-level optimization, solving MaxS-EFT on a training set while solving MaxA-EFT on a validation set, and finds the optimal learning rate and epoch number \textit{without performing iterative gradient descent at all}. This is done by formulating and solving the \textit{Constrained MaxA-EFT},  a variation of MaxA-EFT where $\mDelta$ is constrained to be of the form obtained from gradient descent optimization on loss $\mathcal{L}$ that solves MaxS-EFT. That is Constrained MaxA-EFT constrains MaxA-EFT so that any solution must be obtainable by running gradient descent on $\mathcal{L}$ for some learning rate and number of training steps. This is described next.

%\name{} splits the training set into two and solves MaxA-EFT on one while solving MaxS-EFT on the other in a bilevel optimization. This acts as a regularization, avoiding overfitting in each instance. First, we solve MaxS-EFT using gradient descent, and then solve MaxA-EFT assuming $\vdelta_i$'s are of a form obtained from such a gradient descent optimization. This has two interpretations: (1) MaxA-EFT is solved to set the hyperparameters of gradient descent (i.e., learning rate and number of epochs) performed for MaxA-EFT or (2) gradient descent on MaxS-EFT is used to create a constrained feasible region in MaxA-EFT. Following interpretation (1), we call the splits of training sets used to solve MaxS-EFT and MaxA-EFT training and validation sets, respectively, and denote them by $Q^T, Y^T$ and $Q^V, Y^V$. Nonetheless, interpretation (2) is closer to \name{}'s implementation, where \textit{iterative gradient descent is never performed}, but instead, the gradients are used to reformulate and solve MaxA-EFT.
\textbf{Constrained MaxA-EFT Formulation}. First, consider solving MaxS-EFT with gradient descent on the loss $\mathcal{L}$ in Eq.~\ref{eq:loss} with initialization at zero, i.e., the standard
\begin{align}\label{eq:gd_rec}
    \mDelta^{(0)}\leftarrow \mathbf{0}, \;\mDelta^{(t)}\leftarrow\mDelta^{(t-1)}-\gamma\nabla_{\mDelta}\mathcal{L},
\end{align} when running gradient descent for $t$ iterations with learning rate $\gamma$. Define $\mG=\nabla_{\mDelta}\mathcal{L}$, and observe that  $\mG_i= -\sum_{j\in[n_T]}\mathcal{I}_{i=Y_j^T}\mQ^T_j$, where $\mathcal{I}_{i=Y_j^T}$ is an indicator variable. $\mG$ is independent of $\mDelta$ so that the gradient of $\mathcal{L}$ is constant throughout gradient descent. Using this fact, solving the recursion in Eq.~\ref{eq:gd_rec} we have % $\vdelta_i^{(0)}=0$ for all $i$, and after $t$ iteration of gradient descent with learning rate $\gamma$, we have 
\begin{align}\label{eq:gd:update}
    \mDelta^{(t)}=-\gamma t \mG.
\end{align}

Thus, performing gradient descent on the training set while maximizing validation accuracy is equivalent to solving MaxA-EFT on the validation set assuming $\mDelta$ are of the form of Eq.~\ref{eq:gd:update}. This is formulated as \textit{Constrained MaxA-EFT} problem, setting $\mDelta=\alpha\mG$ in MaxA-EFT, for a scalar parameter $\alpha\geq 0$:
\begin{problem}[Constrained MaxA-EFT]
Let $\mathcal{I}_i(\alpha)$ be an indicator variable denoting whether the $i$-th query is answered correctly after modifying data embeddings by $\alpha \mG$, i.e., when the inequality 
\begin{align}\label{eq:corr_ans_alpha}
    \mQ^V_i\cdot(\mD_{Y_i^V}+\alpha \mG_{Y_i^V}) > \max_{j\in[n]\setminus Y_i^V}\mQ^V_i\cdot(\mD_{j}+\alpha \mG_{j})
\end{align}
    holds. Constrained MaxA-EFT is the optimization problem $\max_{\alpha\geq 0}\sum_{i\in[n_T]}\mathcal{I}_{i}(\alpha)$.
\end{problem}
\fi
\section{Experiments}\label{sec:exp}
We present results on standard text and image retrieval benchmarks and multiple pre-trained models. We present our main experimental results here, but for the sake of space, defer more detailed results and ablation studies to Appx.~\ref{appx:exp}.

\textbf{Datasets}. For text retrieval datasets we use 7 standard datasets: SciFacts \citep{Wadden2020FactOF}, Fever \citep{feverFactExtraction}, ArguAna \citep{argumentationDatax2013} (we use their BEIR \citep{thakur2021beir} versions), TriviaQA~\cite{joshi2017triviaqa}, HotpotQA~\citep{yang2018hotpotqa}, and Natural Questions\citep{kwiatkowski2019natural} (we use their KILT \citep{petroni-etal-2021-kilt} versions), and NF-Corpus \citep{boteva2016} (although all datasets have a BEIR version, we use non-BEIR versions whenever that is larger). We use the datasets as is, without any preprocessing step, except for datasets from KILT, where we only use Wikipedia pages that contain an answer to at least one query (i.e., pages where we expect fine-tuning to have an impact). For image retrieval, we use COCO~\citep{lin2014microsoft} (we use the dataset from 2014) and Flickr~\citep{young2014image} datasets. We use image captions as queries to retrieve the corresponding image. For all text and image datasets, we use 0.7-0.1-0.2 train, validation and test split, but limit test and validation sizes to at most 10,000 queries if there is more. %We create 3 random splits, and all results are averaged across the three runs. 
Statistics about data and query size are presented in Table~\ref{tab:dataset_char}. 
%We specifically use XX taken from BEIR benchmark, XXX taken from KILT benchmark and XX and XX which are standard image retrieval benchmarks.

%For text retrievals, we present results using 

\textbf{Pre-Trained Models}. We report results on fine-tuning embeddings for 5 different pre-trained models, 3 for text and 2 for image retrieval, where we consider models of different sizes and embedding dimensions. For text, we use BGE-small \citep{bge_embedding} with 33M parameters and embedding dimension 384, GTE-large \citep{li2023towards} with 434M parameters and embedding dimension 1024 and OpenAI's text-embedding-large-3 \citep{opeaiemb}, a closed-source model with embedding dimensions 3072. The three models are respectively referred to as BGE-S, GTE-L and TE3-L. %At the time of this writing, BGE-S is one of the most downloaded embedding models on huggingface, GTE-L is best performing under 1billion parameter model on MTEB-retrieval benchmark, and closed-source. 
For image retrieval, we use two CLIP variants \citep{radford2021learning}, ViT-B/32 and ViT-L/14, which have, respectively, 151M and 427M parameters and 512 and 768 embedding dimensions. We respectively call them CLIP-B and CLIP-L.

\textbf{Baselines}. We report results using the embeddings without fine-tuning, called No Fine-Tuning, in addition to training Adaptors and fine-tuning the pre-trained model, referred to as PTFT (see Appx.~\ref{appx:exp:baselines_detail} for implementation details). Due to computational constraints, we report results for the latter only for the small open-source model BGE-S in our main experiments. For both, we present two versions. By default, we use the Multiple Negative Ranking (MNR) loss \citep{henderson2017efficient} for training, which is the standard contrastive fine-tuning loss when positive query/answer pairs are available, suggested by  SentenceTransformers \citep{sbertLossesx2014} and used by LlamaIndex \citep{llamaindex_loss} for fine-tuning (although LlamaIndex only uses a single positive example per query~\citep{llamaindex_loss_labels}). Despite our hyperparameter tuning effort, we observed no accuracy improvements on some datasets through fine-tuning with this loss (see Table~\ref{tab:res_bge_per_dataset}). We then modified the loss, so that only negative samples whose cosine similarity is at least equal to a threshold are included in the loss (see \ref{appx:exp:baselines_detail} for details), which improved accuracy on some datasets but worsened it on other. We use the suffix -L to denote the baselines using this modified loss. We report results using this loss to gain more insight into the baseline's behavior. We use cosine similarity as the retrieval distance metric for No Fine-Tuning, Adaptor, and PTFT.

\textbf{Metrics}. We report typical metrics Recall@$k$ (R@$k$ for short) and NDCG@$k$ with $k=10$ by default, following MTEB benchmark~\citep{muennighoff2022mteb}. We also report Recall@1, the validation metric, and the change in NDCG@$k$ compared with No Fine-Tuning in parenthesis (\textcolor{cadmiumgreen}{+}/\textcolor{burgundy}{-}).

\if 0
\begin{table}[t]
    \centering
    \hspace{-0.5cm}
    \begin{minipage}{0.49\textwidth}
    \centering
        \begin{tabular}{c c c c}
        \toprule
       \textbf{Method}  &  \textbf{R@1} &  \textbf{R@10}&  \textbf{NDCG@10} \\\midrule
\name{} & \textbf{53.3}& \textbf{73.0}& \textbf{61.3} \\
\name{}-N & 46.6& 70.8& 57.4 \\
Adaptor & 45.3& 68.0& 55.3 \\
No Fine-Tuning & 42.3& 67.2& 0.53.5 \\
        \bottomrule
\end{tabular}
    \caption{Results on GTE-large, text datasets avg.}
    \label{tab:res_gte_avg}
    \end{minipage}
    \hspace{0.5cm}
    \begin{minipage}{0.49\textwidth}
    \centering
    \begin{tabular}{c c c c}
        \toprule
       \textbf{Method}  &  \textbf{R@1} &  \textbf{R@10}&  \textbf{NDCG@10} \\\midrule
\name{} & \textbf{50.6}& 66.9& \textbf{57.4} \\
\name{}-N & 45.1& \textbf{68.7}& 55.6 \\
Adaptor & 38.6& 63.4& 49.8 \\
No Fine-Tuning & 37.7& 62.3& 48.7 \\
        \bottomrule
\end{tabular}
    \caption{Results on BGE-small, text datasets avg.}
    \label{tab:res_gte_avg}
    \end{minipage}
        \begin{minipage}{0.49\textwidth}
    \centering
    \begin{tabular}{c c c c}
        \toprule
       \textbf{Method}  &  \textbf{R@1} &  \textbf{R@10}&  \textbf{NDCG@10} \\\midrule
\name{} & 54.4& 75.1& 62.8 \\
\name{}-N & 52.4& 75.4& 62.2 \\
Adaptor & 47.0& 66.8& 54.9 \\
No Fine-Tuning & 40.0& 67.6& 52.2 \\
        \bottomrule
\end{tabular}
    \caption{Results on BGE-small, text datasets avg.}
    \label{tab:res_gte_avg}
    \end{minipage}
\end{table}

\fi

\begin{table}[t]
\hspace{-2cm}
        \begin{tabular}{c c c c c c c c c c}
        \toprule
       &\textbf{NF-Corpus}&\textbf{SciFact}&\textbf{ArguAna}&\textbf{Fever}&\textbf{NQ}&\textbf{TriviaQA}&\textbf{HotpotQA}&\textbf{COCO}&\textbf{Flickr}
   \\\midrule
\textbf{Query \#}&2,429&1,109&1,401&123,142&79,782&58,245&74,259&414,113&155,070\\
\textbf{Data \#}&3,633&5,183&8,674&5,416,568&7,631,395&7,631,395&7,631,395&82,783&31,014\\
\bottomrule
\end{tabular}
    \caption{Total number of queries and records in the datasets}
    \label{tab:dataset_char}
\end{table}

%\name{} & \textbf{50.6}& 66.9& \textbf{57.4}& \textbf{53.3}& \textbf{73.0}& \textbf{61.3} & \textbf{54.4}& \textbf{75.1}& \textbf{62.8} \\
%\name{}-N &  45.1& \textbf{68.7}& 55.6 &46.6& 70.8& 57.4 & 52.4& \textbf{75.4}& 62.2 \\

\if 0
\begin{table}[t]
\hspace{-1.5cm}
        \begin{tabular}{c | c c c | c c c | c c c}
        \toprule
      \textbf{Emb. Model \textrightarrow} & \multicolumn{3}{c}{\textbf{BGE-S}}& \multicolumn{3}{|c}{\textbf{GTE-L}}
         & \multicolumn{3}{|c}{\textbf{TE3-L}}
         \\
        \textbf{\textdownarrow \; Method} &  \textbf{R@1} &  \textbf{R@10}&  \textbf{NDCG@10}  &  \textbf{R@1} &  \textbf{R@10}&  \textbf{NDCG@10}  &  \textbf{R@1} &  \textbf{R@10}&  \textbf{NDCG@10} \\\midrule
\name{}-M & 50.5& 66.6& 57.3& 52.6& 72.5& 60.8 & 54.4& 75.3& 63.1 \\
\name{}-N &  \textbf{52.5}& \textbf{72.3}& \textbf{61.1} &\textbf{53.8}& \textbf{75.0}& \textbf{62.9} & \textbf{55.3}& \textbf{75.9}& \textbf{63.9} \\
Linear Adaptor &   40.3& 64.5& 51.2& 45.3& 68.0& 55.3 & 47.4& 66.6& 54.9 \\
2-Layer Adaptor & 38.0& 63.1& 49.3  & 42.6& 67.8& 54.1 
&42.8& 69.2& 54.4 \\
PTFT & 40.7& 65.9& 52.3  & N/A& N/A& N/A & N/A& N/A& N/A\\
No Fine-Tuning & 37.7& 62.3& 48.7& 42.3& 67.2& 53.5& 40.0& 67.6& 52.2  \\
        \bottomrule
\end{tabular}
    \caption{Average results across text datasets grouped by the embedding model used}
    \label{tab:res_text_avg}
\end{table}
\fi

\begin{table}[t]
\hspace{-1.5cm}
        \begin{tabular}{c | c c c | c c c | c c c}
        \toprule
      \textbf{Emb. Model \textrightarrow} & \multicolumn{3}{c}{\textbf{BGE-S}}& \multicolumn{3}{|c}{\textbf{GTE-L}}
         & \multicolumn{3}{|c}{\textbf{TE3-L}}
         \\
        \textbf{\textdownarrow \; Method} &  \textbf{R@1} &  \textbf{R@10}&  \textbf{NDCG@10}  &  \textbf{R@1} &  \textbf{R@10}&  \textbf{NDCG@10}  &  \textbf{R@1} &  \textbf{R@10}&  \textbf{NDCG@10} \\\midrule
\name{}-M & 49.7  & 66.6  & 57.0  \textcolor{cadmiumgreen}{(+8.4)}& 52.1  & 73.4  & 61.0  \textcolor{cadmiumgreen}{(+7.7)}& 54.1  & \textbf{75.6  }& 63.2  \textcolor{cadmiumgreen}{(+11.0)}\\
\name{}-N & \textbf{52.0  }& \textbf{72.6  }& \textbf{61.1  \textcolor{cadmiumgreen}{(+12.4)}}& \textbf{53.4  }& \textbf{74.8  }& \textbf{62.7  \textcolor{cadmiumgreen}{(+9.4)}}& \textbf{55.2  }& \textbf{76.0  }& \textbf{63.9  \textcolor{cadmiumgreen}{(+11.7)}}\\
Adapter & 39.5  & 65.5  & 51.6  \textcolor{cadmiumgreen}{(+2.9)}& 45.1  & 68.4  & 55.7  \textcolor{cadmiumgreen}{(+2.4)}& 46.9  & 66.2  & 54.4  \textcolor{cadmiumgreen}{(+2.2)}\\
PTFT & 40.9  & 66.1  & 52.5  \textcolor{cadmiumgreen}{(+3.8)}& N/A& N/A& N/A& N/A& N/A& N/A\\
No Fine-Tuning & 37.0  & 62.4  & 48.7  & 41.6  & 67.0  & 53.3  & 40.0  & 67.5  & 52.2  \\
        \bottomrule
\end{tabular}
    \caption{Average results across text datasets grouped by the embedding model used}
    \label{tab:res_text_avg}
\end{table}

%\name{} & 28.7& 56.2& 41.6  & 31.6& 59.3& 44.5 \\
%\name{}-N & \textbf{30.1}& \textbf{57.3}& \textbf{42.8}  & \textbf{32.7}& \textbf{60.6}& \textbf{45.7} \\

\begin{table}[t]
\hspace{-2.5cm}
\begin{minipage}{0.7\textwidth}
\centering
        \begin{tabular}{c | c c c | c c c}
        \toprule
      \textbf{Emb. Model \textrightarrow} & \multicolumn{3}{c}{\textbf{CLIP-B}}& \multicolumn{3}{|c}{\textbf{CLIP-L}}
         \\
        \textbf{\textdownarrow \; Method} &  \textbf{R@1} &  \textbf{R@10}&  \textbf{NDCG@10}  &  \textbf{R@1} &  \textbf{R@10}&  \textbf{NDCG@10}   \\\midrule
\name{}-M & \textbf{28.6  }& \textbf{55.1  }& \textbf{40.9  \textcolor{cadmiumgreen}{(+14.1)}}& \textbf{30.1  }& \textbf{58.1  }& \textbf{43.2  \textcolor{cadmiumgreen}{(+10.7)}}\\
\name{}-N & \textbf{28.8  }& \textbf{55.3  }& \textbf{41.1  \textcolor{cadmiumgreen}{(+14.3)}}& \textbf{30.1  }& \textbf{58.2  }& \textbf{43.3  \textcolor{cadmiumgreen}{(+10.8)}}\\
Adapter & 18.5  & 43.5  & 29.9  \textcolor{cadmiumgreen}{(+3.0)}& 24.1  & 50.9  & 36.5  \textcolor{cadmiumgreen}{(+4.0)}\\
No Fine-Tuning & 15.9  & 40.1  & 26.9  & 20.5  & 46.5  & 32.5  \\\bottomrule%Adaptor-L &15.9& 40.2& 26.9  \textcolor{gray}{(+0)}&20.4& 46.5& 32.5 \textcolor{gray}{(+0)}\\
%2-Layer Adaptor &  17.3& 42.3& 28.6 & 22.4& 50.6& 35.5 \\
\end{tabular}
    \caption{\hbox{Average results across image datasets grouped by the embedding model used}}
    \label{tab:res_img_avg}
\end{minipage}
\hspace{2.4cm}
\begin{minipage}{0.35\textwidth}
%\vspace{-0.46cm}
\centering
        \begin{tabular}{>{\centering}p{1.67cm} >{\centering}p{1.6cm} >{\centering\arraybackslash}p{1.58cm}}
        \toprule
      \multirow{2}{*}{\textbf{Method}} & \centering\textbf{Time GPU (mins.)} & \textbf{Time CPU (mins.)}
   \\\midrule
\name{}-M & \textbf{1.14}&\textbf{7.12}\\
\name{}-N & 2.18 &11.0\\
Adaptor & 7.99&77.8\\
PTFT & 447 &N/A\\\bottomrule
\end{tabular}
    \caption{\hbox{Run time using BGE-S}}
    \label{tab:runtime}
\end{minipage}
\end{table}

\begin{table}[t]
\hspace{-2cm}
%\centering
        \begin{tabular}{c c c c c c c c}
        \toprule
       \textbf{Method} &\textbf{ArguAna}&\textbf{Fever}&\textbf{HotpotQA}&\textbf{NF-Corpus}&\textbf{NQ}&\textbf{SciFact}&\textbf{TriviaQA}
   \\\midrule
\name{}-M & \textbf{47.9  \textcolor{gray}{(+0)}}& \textbf{95.0  \textcolor{cadmiumgreen}{(+13.7)}}& 63.1  \textcolor{cadmiumgreen}{(+5.0)}& 40.0  \textcolor{cadmiumgreen}{(+6.1)}& 38.6  \textcolor{cadmiumgreen}{(+17.5)}& 79.2  \textcolor{cadmiumgreen}{(+6.5)}& 35.5  \textcolor{cadmiumgreen}{(+9.6)}\\
\name{}-N & \textbf{47.9  \textcolor{gray}{(+0)}}& 93.5  \textcolor{cadmiumgreen}{(+12.3)}& \textbf{65.2  \textcolor{cadmiumgreen}{(+7.1)}}& 44.6  \textcolor{cadmiumgreen}{(+10.7)}& \textbf{45.9  \textcolor{cadmiumgreen}{(+24.8)}}& \textbf{87.8  \textcolor{cadmiumgreen}{(+15.1)}}& \textbf{43.0  \textcolor{cadmiumgreen}{(+17.2)}}\\
Adapter & \textbf{47.9  \textcolor{gray}{(+0)}}& 81.2  \textcolor{gray}{(+0)}& 58.1  \textcolor{gray}{(+0)}& 37.4  \textcolor{cadmiumgreen}{(+3.5)}& 27.4  \textcolor{cadmiumgreen}{(+6.2)}& 83.5  \textcolor{cadmiumgreen}{(+10.9)}& 25.9  \textcolor{gray}{(+0)}\\
Adapter-L & \textbf{47.9  \textcolor{gray}{(+0)}}& 88.4  \textcolor{cadmiumgreen}{(+7.2)}& 62.7 \textcolor{cadmiumgreen}{(+4.6)}& 35.7  \textcolor{cadmiumgreen}{(+1.9)}& 33.6 \textcolor{cadmiumgreen}{(+12.0)}& 85.1  \textcolor{cadmiumgreen}{(+12.4)}& 27.2 \textcolor{cadmiumgreen}{(+1.2)}\\
PTFT & \textbf{47.9  \textcolor{gray}{(+0)}}& 81.2  \textcolor{gray}{(+0)}& 58.1  \textcolor{gray}{(+0)}& \textbf{46.1  \textcolor{cadmiumgreen}{(+12.2)}}& 28.1  \textcolor{cadmiumgreen}{(+7.0)}& 80.4  \textcolor{cadmiumgreen}{(+7.7)}& 25.9  \textcolor{gray}{(+0)}\\
PTFT-L & \textbf{47.9  \textcolor{gray}{(+0)}}& 84.1  \textcolor{cadmiumgreen}{(+2.9)}& 62.1  \textcolor{cadmiumgreen}{(+4.0)}& 36.0  \textcolor{cadmiumgreen}{(+2.2)}& 36.8  \textcolor{cadmiumgreen}{(+15.6)}& 72.7  \textcolor{gray}{(+0)}& 26.0  \textcolor{cadmiumgreen}{(+0.1)}\\
No Fine-Tuning & \textbf{47.9  }& 81.2  & 58.1  & 33.8  & 21.2  & 72.7  & 25.9  \\
\bottomrule
\end{tabular}
    \caption{NDCG@10 results for using BGE-S on text datasets}
    \label{tab:res_bge_per_dataset}
\end{table}

\begin{table}[t]
\hspace{-1.5cm}
\begin{minipage}{0.75\textwidth}
%\hspace{-2cm}
%\begin{minipage}{0.7\textwidth}
    \centering
    %\hspace{-0.5cm}
    \centering
        \begin{tabular}{c | c c c | c c c}
        \toprule
       \multirow{2}{*}{\textbf{Method}}& \multicolumn{3}{c}{\textbf{In-Distribution}}& \multicolumn{3}{|c}{\textbf{Out-of-Distribution}}\\%\cline{2-7}
         &  \textbf{R@1} &  \textbf{R@10}&  \textbf{NDCG@10} &  \textbf{R@1} &  \textbf{R@10}&  \textbf{NDCG@10} \\\midrule
\name{}-M & 49.5& 65.5& 56.3 \textcolor{cadmiumgreen}{(+9.4)}& 30.8& 46.9& 38.0  \textcolor{burgundy}{(-10.0)}\\%\hline
\name{}-N & \textbf{51.1}& \textbf{71.5}& \textbf{60.0  \textcolor{cadmiumgreen}{(+13.1)}}& \textbf{40.7}& \textbf{64.3}& \textbf{51.2 \textcolor{cadmiumgreen}{(+3.2)}} \\%\hline
Adaptor & 40.1&58.6&47.6 \textcolor{cadmiumgreen}{(+0.7)}&37.3&	56.6&	45.0 \textcolor{burgundy}{(-3.0)} \\%\hline
No Fine-Tuning & 37.7& 59.1& 46.9 & 39.5& 60.1& 48.0 \\\bottomrule
\end{tabular}
    \caption{Distribution shift results using BGE-S, average over text datasets}
    \label{tab:dist_shift}
    %\end{minipage}
\end{minipage}
\hspace{1.6cm}
\begin{minipage}{0.35\textwidth}
%\begin{minipage}{0.3\textwidth}
\centering
        \begin{tabular}{>{\centering}p{2.3cm} c}
        \toprule
      \textbf{Method} & \textbf{NDCG@10}
   \\\midrule
\name{}-N & \textbf{61.1}\\
\name{}-M+N & 58.8\\
\name{}-M & 57.0\\
%\name{}-IN & \textbf{61.8} \\
\name{}-NU & 56.6\\
No Fine-Tuning & 48.7\\
%\name{}-IM+R & 52.4
\bottomrule
\end{tabular}
    \captionof{table}{Ablation of \name{}}
    \label{tab:norm_ablation_mainbody}
%\end{minipage}
\end{minipage}
\end{table}

%Optimal GD & 41.7& 64.2& 51.5 & \textbf{3.53}\\

\subsection{Baseline Results}\label{sec:exp:baseline}
\textbf{Summary of results}. Tables~\ref{tab:res_text_avg}-\ref{tab:res_img_avg} present our accuracy results averaged across all text or image datasets for different models.  Table~\ref{tab:res_bge_per_dataset} presents the per dataset results for BGE-S. The per dataset results for other models followed similar trends and are deferred to Appx.~\ref{appx:per_dataset}.  

As Tables~\ref{tab:res_text_avg}-\ref{tab:res_img_avg} show, both \name{}-M and \name{}-N provide significant accuracy gains, providing up to 14.3\% NDCG@10 boost over No Fine-Tuning while PTFT and Adaptor only improve NDCG@10 up to 4.0\%, when averaged across datasets, for both text and image retrieval. Interestingly, the accuracy gains from \name{}-M and \name{}-N depends on the pre-trained model. GTE-L outperforms TE3-L without fine-tuning, but using \name{} TE3-L outperforms GTE-L. %Thus, differences in the embedding space can impact the performance of a fine-tuning approach differently.  

Moreover, Table~\ref{tab:runtime} shows the total fine-tuning time to run BGE-S on our text datasets (i.e., time obtain the associated results in Table~\ref{tab:res_text_avg}) using an Nvidia A100 GPU as well as using 32 core Intel Broadwell CPUs. The reported time excludes the time to embed the data records (which is the same across all methods). \name{} variants run in 1-2 minutes with GPU and in up to 11 minutes using CPUs, which is, respectively, more than 3 and 11 times faster than Adaptor and more than 200 times faster than PTFT, which cannot be run on CPUs in a reasonable time-frame. For both Adaptor and PTFT, the reported run times are for optimized implementations that include early stopping and optimizations for efficient calculation of validation accuracy (see Appx.~\ref{appx:exp:baselines_detail}).

\textbf{Detailed Results}. Table~\ref{tab:res_bge_per_dataset} shows the detailed retrieval accuracy on text datasets for BGE-S. \name{} significantly outperforms parametric methods on almost all datasets. \name{}-N outperforms \name{}-M on most datasets, showing the benefit of constraining embeddings to be normalized. 

The results show that parametric approaches are unreliable and fail to provide significant accuracy improvements despite requiring significantly more computational resources. %, both for the fine-tuning itself and hyperparameter tuning. % for fine-tuning  as well as hyperparameter tuning, they improve accuracy on some datasets (although never as much as \name{}), but fail to provide any improvement on others (the training loss goes down on all dataset, but validation accuracy only decreases on some, resulting in a final model the same as No Fine-Tuning on those datasets). 
The reported results are after hyperparameter tuning as well as tweaking the loss function (i.e., -L variants). We observed that the latter does help improve (and sometimes worsen) accuracy for parametric approaches on some datasets, with Adaptor-L and PTFT-L providing accuracy improvements on Fever and HotpotQA where Adaptor and PTFT provided none. %(Adaptor-L does worse than No Fine-Tuning on ArguAna in terms of NDCG@10, but does better in terms of R@1, and the latter was used as validation metric). 
Meanwhile, \name{} consistently provides significant accuracy boosts \textit{without any hyperparameter tuning}. We provide a detailed discussion of the failure modes of the parametric approaches in Appx.~\ref{sec:accuracy_during_training}. 

To better understand the results, we remark on the performance on ArguAna and Fever datasets. Fever, where \name{}-M performs better than \name{}-N has a skewed label distribution, where the same paragraph is the correct answer for many queries. This allows for setting the magnitude of the embeddings based on label distribution, assigning embeddings with larger magnitudes to more frequently accessed passages. Such an assignment can improve the accuracy when the label distribution is fixed and skewed, but leads to worse generalization when there is no skew (as the results on other datasets in Table~\ref{tab:res_bge_per_dataset} show) or when there is a distribution shift (see Sec.~\ref{sec:exp:dist_shift}). Finally, in ArguAna, each data record is a factual argument and each query provides an argument and asks for a counterargument to the given argument. In such a setting, to improve accuracy, we expect larger systematic changes to the embedding space to be required to be able to retrieve the semantically opposite (instead of similar). Learning such changes from a small training set is challenging %, and ArguAna represents a difficult fine-tuning scenario in which none of the approaches performs well, 
and perhaps a more task-specific methodology is required for this dataset.

%We do not expect fine-tuning with semantic similarity, as is, to improve accuracy since the task asks for the opposite document to a query.

%. Here, we briefly note that beside ArguAna where no approach improves accuracy (this could be due to XXXXXXXX), fine-tuning with Adaptors and PTFT doest not improve accuracy for none of Fever, HotpotQA and TriviaQA datasets. In fact, even though 

%While we have made standard choices for the loss function of parametric approaches, and have performed hyperparameter tunning to set their parameters, we note that it may be possible to improve the accuracy for parametric approaches through spending significantly more engineering effort and computational resources, and by performing per dataset tweaking of the loss and/or additional data processing. 
%even though for both Adapter and PTFT approaches we observed that fine-tuning reduces the training loss on all dataset, the validation and training \textit{accuracy} did not improve on these datasets. 

%Moreover, we see that on text datasets, \name{} outperforms \name{}-N, while the order is reversed on image datasets. Given that both the embedding models and the datasets are different between Tables~\ref{tab:res_text_avg} and \ref{tab:res_img_avg}. 

\subsection{Out-of-distribution Generalization}\label{sec:exp:dist_shift}
%Results in Sec.~\ref{sec:exp:baseline} show the performance of models on in-distribution queries. 
Next, we study the impact of fine-tuning on out-of-distribution queries. In this experiment, separately for each dataset, we use K-means to cluster all the queries in two clusters, referred to as $C_1$ and $C_2$. %, respectively used for training and in-distribution training and $C_2$ for out-of-distribution testing. 
We split $C_1$ into 3 sets, $C_1^{\text{train}}$, $C_1^{\text{val}}$ and $C_1^{\text{test}}$, where  $C_1^{\text{train}}$ and $C_1^{\text{val}}$ are used for training and validation. We report test results on $C_1^{\text{test}}$ as in-distribution and on $C_2$ as out-of-distribution results. 

Average results across text datasets and with BGE-S model are shown in Table~\ref{tab:dist_shift}. As the table shows, \name{}-N performs the best on the out-of-distribution test set, even outperforming No Fine-Tuning, while providing a significant accuracy boost on the in-distribution samples. Although \name{}-M performs well on in-distribution samples, its performance deteriorates on out-of-distribution queries. The main difference between \name{}-M and \name{}-N is that \name{}-M's embeddings are not normalized. Thus, when retrieving top-$k$ results using inner product as similarity metric, a fine-tuned embedding with large magnitude can adversely impact the query answers for out-of-distribution queries. However, by keeping embeddings normalized, \name{}-N ensures fine-tuned embeddings do not change the answer to queries that are far from fine-tuned data records, thus avoiding performance degradation on out-of-distribution queries. Finally, Adaptor provides little gain on in-distribution queries, while worsening accuracy on out-of-distribution samples. 

\if 0
\begin{figure}
\hspace{-0.7cm}\begin{minipage}{0.7\textwidth}
    \centering
    \includegraphics[width=1\linewidth]{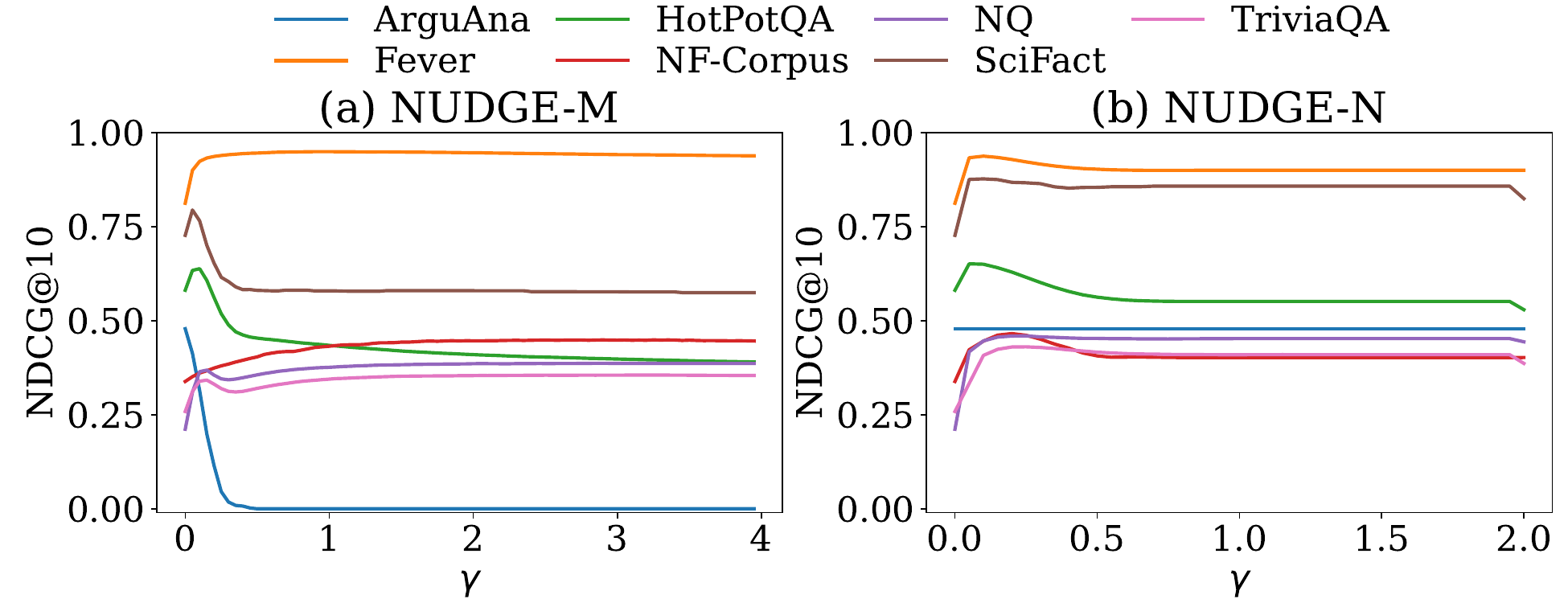}
    \caption{Impact of $\gamma$ using BGE-S}
    \label{fig:exp:gamma}
\end{minipage}
\hspace{0.1cm}
\begin{minipage}{0.35\textwidth}
%\begin{minipage}{0.3\textwidth}
\centering
        \begin{tabular}{>{\centering}p{2.2cm} c}
        \toprule
      \textbf{Method} & \textbf{NDCG@10}
   \\\midrule
\name{}-N & \textbf{61.1}\\
%\name{}-IN & \textbf{61.8} \\
\name{}-M+N & 58.8\\
%\name{}-IM+R & 52.4
\bottomrule
\end{tabular}
    \captionof{table}{Normalization study, avg. accuracy using BGE-S}
    \label{tab:norm_ablation_mainbody}
%\end{minipage}
\end{minipage}
\end{figure}
\fi

\subsection{Ablation Study}\label{sec:exp:ablation}
We provide an ablation study to better understand the impact of various constraints in \name{}. In addition to \name{}-M and \name{}-N, we present \name{}-M+N, which performs normalization on the output of \name{}-M. That is, instead of incorporating normalization as a constraint in the optimization problem, \name{}-M+N simply normalizes the embeddings after performing \name{}-M. We also present \name{}-NU, which is a variation of \name{}-N that only includes the normalization constraint, but not the constraint on the magnitude of the change in the embeddings (the solution is equivalent to the first branch of Eq.~\ref{eq:nopetn_delta_gamma}). 

Table~\ref{tab:norm_ablation_mainbody} shows the results of comparing the above variants across text datasets using the BGE-S model. The table shows normalizing the embeddings as part of the optimization (\name{}-N) is better than simply normalizing the output after optimization (\name{}-M+N), although both work better than using unnormalized embeddings (\name{}-M). Moreover, allowing the magnitude of change to the embeddings to be unbounded (\name{}-NU) performs worse than all other variants, showing the benefit of constraining how much embeddings can change during fine-tuning. Overall, comparing \name{}-N, No Fine-Tuning, and all other variants, we see that the benefits of \name{}-N come from a combination of all design choices made. 

%Table~\ref{tab:norm_ablation_mainbody} compares \name{}-N and , denoted by  Results show that, indeed, it is better to include normalization as a constraint within optimization, compared with normalizing the output after optimization. See Sec.~\ref{exp:norm:ablation} for more experimental results on normalization.

\if 0
\textbf{Over-fitting to Label Distribution}
For text datasets only use test/train samples with the same number of labels?

\subsection{New Document Insertion}
\fi

%\subsection{Scalability}

\if 0
Scalability exps
Impact of positive/negative examples -> classification loss vs contrastive loss
Training is finicky?
Hyperparameter tuning?

Run llm fine tuning for the larger datasets

Compare runtime for larger embeddings?

Need to do a dist shift exp

Show exp when train model to mimic kNN doesn't work

Fine tune specific for retrieval

- New Documents?!
Should do exps with adding docs?
- Docs that are answer?
- Docs that aren't answer
- Maybe ADD A PAR IF ITS NOT TOO BAD

- Does embedding performantnce become worse for other tasks?

\fi
\section{Related Work}
Decades of research has explored various retrieval methods, including sparse
\citep{robertson2009probabilistic}, dense \citep{karpukhin2020dense}, late interaction \cite{khattab2020colbert} and generative retrieval \cite{tay2022transformer} to name a few. Compared with such retrieval paradigms, $k$-NN retrieval has recently gained increased popularity (see \cite{zhao2024dense} for a survey of recent studies) due to its simplicity and efficiency. $k$-NN retrieval uses embeddings from pre-trained models, thus avoiding the need for extensive training on the dataset in hand, and performs retrieval through a simple vector index look-up avoiding expensive model inference at query time. 

This paper focuses on improving $k$-NN retrieval accuracy given access to a pre-trained embedding model. Related work can be divided into two categories. The first category, including this paper, aims to improve the embeddings through fine-tuning for the specific dataset and query workload. Fine-tuning the pre-trained model itself, through a similar training strategy used during pre-training is possible, \cite{zhao2024dense, tang2022dptdr, tam2022parameter, liu2021p, pires2019multilingual, yu2022coco,ma2024fine}, but requires access to the model parameters, is computationally expensive and incurs extra hosting and maintenance costs the fine-tuned models at deployment time. A common alternative in practice is to train adaptors \citep{trychromaEmbeddingAdapters, llamaindexFineTuningLlamaIndex} to transform the output of the models. As our experiments show, adaptors provide limited accuracy gains. On the other hand, our approach, \name{}, is efficient, model agnostic, does not require hosting and maintaining any model at deployment time, and provides a significant accuracy boost. Work in the second category, orthogonal to this paper, modify the retrieval algorithm to improve accuracy, e.g., through query rewriting, reranking, etc. (see \cite{zhao2024dense, gao2023retrieval} for a survey). Notable examples include \cite{gao2022precise}, which rewrites the queries as a hypothetical documents, \cite{sarthi2024raptor}, which adds summary text chunks to the dataset to improve retrieval, \cite{sachan2022improving} which reranks passages based on probability of observing the query, and \cite{lin2024towards} which extracts document structure for query answering. 

Beyond retrieval, our work is related to recent work on fine-tuning pre-trained models for various purposes~\citep{ouyang2022training, rafailov2024direct, ziegler2019fine, zhang2024raft, patil2023gorilla}, but shows the novel insight that non-parametric fine-tuning of model output, instead of fine-tuning model parameters, can provide significant accuracy improvements. Similar to \cite{ouyang2022training, rafailov2024direct, ziegler2019fine}, our optimization constrains how much model output can change during fine-tuning. In \name{}, this constraint represents the intuition that among all fine-tuned embeddings that achieve a specific training loss, one with the least change from non-fine-tuned embedding should be preferred. Although this intuition can be broadly applicable to fine-tuning, the KL-divergence constraint in \cite{ouyang2022training, ziegler2019fine} has additional significance due to the use of a reward model.

%Finally, we note in passing that although our approach can be used to improve the retrieval component in RAG piplines~\citep{lewis2020retrieval}, it is orthogonal to recent efforts in improving the RAG system as whole, e.g., by metadata extraction~\citep{edge2024local}, or fine-tuning the generative component~\citep{zhang2024raft}.

\if 0
\textbf{Retrieval}. 

-  A lot of work on retrieval 
-  Nearest neighbor retrieval with pretrained embedding has become very common

\textbf{DSI, Generative. retrieval}

\textbf{Improving Retreival with Pre-Trained Models}
- Hyde, reranking, etc
- orthogonal to our work

\textbf{Fine Tuning Pre-Trained Models}. 
- A lot of work fine tune for applications
- A lot of work also on scalable fine tuning
- Fine tuning for RAG, Raft
- Differences
\fi
\section{Conclusion}
We studied the problem of fine-tuning embeddings to improve $k$-NN retrieval accuracy. We presented \name{}, a novel non-parametric approach to embedding fine-tuning that is efficient, provides significant accuracy boosts and does not require any model hosting or maintenance at deployment time. %\name{} is non-parametric, that is, unlike existing work that fine-tunes model parameters to improve the embeddings, \name{} modifies the embeddings themselves. 
Our experimental results show \name{} improves accuracy by up to 16.0\% compared with existing methods and up to 24.4\% compared with no fine-tuning. Future work includes incorporating \name{} inside vector databases and generating and maintaining query sets for fine-tuning.

%\subsubsection*{Acknowledgments}
%Use unnumbered third level headings for the acknowledgments. All
%acknowledgments, including those to funding agencies, go at the end of the paper.

\bibliography{iclr2024_conference}
\bibliographystyle{iclr2024_conference}

%\appendix
%\section{Appendix}
%You may include other additional sections here.
\appendix
\section{Appendix Overview}
This appendix is organized as follows:
\begin{itemize}
    \item Appx.~\ref{appx:proofs} contains the proofs of the theoretical results in the paper. 
    \item Appx.~\ref{sec:method:practice} contains other practical \name{} variants. 
    \item Appx.~\ref{appx:multi_label} discusses the extension of \name{} to multi-label settings. 
    \item Appx.~\ref{appx:exp} provides details about the experimental setting, and provides additional experiments. 
\end{itemize}
\section{Proofs}\label{appx:proofs}
Appx.~\ref{appx:np_hard_proof}-\ref{proof:them:nopetn}, respectively, provide the proofs of Theorems.~\ref{thm:np_hard}-\ref{thm:nopet_n}. The proofs of technical lemmas are presented in Appx.~\ref{appx:lemmas}

\subsection{Proof of Theorem~\ref{thm:np_hard}}\label{appx:np_hard_proof}
We show a reduction from \textit{homogeneous maximum feasible linear subsystem, Max-FLS,} shown to be NP-Hard in \cite{amaldi1995complexity}. We first formally state the decision version of the problem.

\begin{problem}[Homogeneous Max-FLS]
    Given a linear system $\mA\vx< \mathbf{0}$, where $\mA$ is of size $s\times t$, and an integer $K$ with $1 <K < s$, does there exist a solution $\vx\in\mathcal{R}^{t}$ satisfying at least $K$ inequalities of the system?
\end{problem}

We show a reduction from Homogeneous Max-FLS to the decision version of MaxA-EFT. The decision version of MaxA-EFT asks whether there exists $\mDelta$ such that the training accuracy is at least $K$, i.e., whether at least $K$ different $\mathcal{I}_i(\mDelta)$, $i\in[n_T]$ can be satisfied.

Given $\mA$, the reduction defines an instance of MaxA-EFT by specifying $\mD$, $\mQ$, and $Y$. Specifically, we let $\mQ=\mA$, $\mD=\mathbf{0}\in\mathcal{R}^{2\times t}$ (i.e., a $2\times t$ matrix only consisting of zeros) and $Y=\{Y_1, ..., Y_s\}$, $Y_i=0\;\forall i\in[s]$ (i.e., this instance only has 2 data records, embedding dimension is $t$ and there are $s$ training queries). 

We next show that there exists $\mDelta$ so that the training accuracy of MaxA-EFT is at least $K$ if and only if there exists a solution $\vx$ to Homogeneous Max-FLS satisfying at least $K$ inequalities of the system. 

Observe that, by the above construction, $\mathcal{I}_i(\mDelta)$ is satisfied if and only if 
$$
\mQ_i\cdot (\mDelta_1-\mDelta_0)< 0. 
$$
\if 0
or equivalently the $i$-th inequality
$$
\mQ (\mDelta_1-\mDelta_0)< \mathbf{0}. 
$$
\fi
Suppose the training accuracy is at least $K$ for some $\mDelta$. Let $\vx=\mDelta_1-\mDelta_0$, and let $\{i_1, ..., i_K\}$ be the index of $K$ training samples answered correctly. Thus, we have $\mA_i\cdot\vx< 0$ for all $i\in \{i_1, ..., i_K\}$, showing the existing $K$ inequalities in the subsystem that are satisfied. 

Conversely, assume some $\vx$ satisfies at least $K$ inequalities in the subsystem. Let $\mDelta_0=\mathbf{0}$ and $\mDelta_1=\vx$. We have that at least $K$ inequalities in
$$
\mQ \mDelta_1< \mathbf{0}
$$
are satisfied, so that the training accuracy is at least $K$. The reduction is polynomial time, thus showing MaxA-EFT is NP-hard.\qed

\subsection{Proof of Theorem~\ref{thm:nopet}}\label{proof:them:nopetm}
\begin{algorithm}[t]
\begin{algorithmic}[1]
\Require A training set $\mQ^T, Y^T$, validation set $\mQ^V, Y^V$ and data embeddings $\mD$
\Ensure Fine-tuned data embeddings 
\State $\mG_i\leftarrow \sum_{j\in[n_T]}\mathds{I}[i=Y_j^T]\mQ^T_j$ for all $i\in [n]$\label{alg:line:grad}
\State Calculate $\mathcal{G}_{i, j}, \mathcal{S}_{i, j}$ for all $i\in [n_V], j\in [n]$
\State $I\leftarrow \emptyset$
\For{$i$ \textbf{in} $[n_V]$}\label{alg:line:for_qv}
    \If {$\mathcal{S}_{i, j}>\mathcal{S}_{i, Y_i^V}$ for any $j\in \{j\in[n]\setminus Y_i^V|\QG{}=0\}$}\label{alg:line:check_neg}
        \State \textbf{Continue}\Comment{$I_i=\emptyset$ so skip the query}
    \EndIf 
    \State $P\leftarrow\{j\in[n]\setminus Y_i^V|\QG{}>0\}$\label{alg:line:p}
    \State $N\leftarrow\{j\in[n]\setminus Y_i^V|\QG{}<0\}$
    \State $l\leftarrow \max_{j\in P}\frac{\mathcal{S}_{i, j}-\mathcal{S}_{i, Y_i^V}}{\QG{}}$, $u\leftarrow \min_{j\in N}\frac{\mathcal{S}_{i, j}-\mathcal{S}_{i, Y_i^V}}{\QG{}}$\label{alg:line:l_u}\label{alg:line:bounds}
    \If{$u<0$ \textbf{or} $u<l$}
        \State \textbf{Continue}\Comment{$I_i=\emptyset$ so skip the query}
    \EndIf
    \State $I_i\leftarrow (\max\{l, 0\}, u)$
    \State $I.$append($I_i$)\label{alg:line:add_R}
\EndFor
\State $\gamma^*\leftarrow\arg\max_{\gamma}\sum_{I_i\in I}\mathds{I}[\gamma\in I_i]$\label{alg:line:final_gamma}
\State $\mD_i^*\leftarrow\mD_i+\gamma^* \mathds{I}[\normx{\mG_{i}}\neq0]\frac{\mG_i}{\normx{\mG_i}}$
\State \textbf{return} $\mD^*$
\caption{\name{}-M algorithm}\label{alg:optimval_val}
\end{algorithmic}
\end{algorithm}

\textbf{Setup}. Recall that for any $i\in[n]$, 
$$\mG_i= \sum_{j\in[n_T]}\mathds{I}[i=Y_j^T]\mQ^T_j,$$
so that 
$$
\underset{\mDelta\in\mathcal{R}^{n\times d}}{\arg\max} \sum_{i\in[n_T]}\mQ^T_i\cdot (\mD_{Y_i^T}+\mDelta_{Y_i^T})=\underset{\mDelta\in\mathcal{R}^{n\times d}}{\arg\max} \sum_{i\in[n]}\mG_i\cdot \mDelta_{i}=\underset{\mDelta\in\mathcal{R}^{n\times d}}{\arg\max} \sum_{i\in[n], \normx{\mG_i}\neq0}\mG_i\cdot \mDelta_{i}.
$$
The proofs here use the above formulation for \texttt{MaxS-M}. Moreover, we set $\mDelta_i^\gamma=\textbf{0}$ for any $i$ where $\normx{\mG_i}=0$, given that they do not appear in the above objective. %and exclude such $i$ values from the optimization when finding \texttt{MaxS-M}, so that the objective is 
\if 0
$$
\underset{\mDelta\in\mathcal{R}^{n\times d}}{\arg\max} \sum_{i\in[n], \normx{\mG_i}\neq0}\mG_i\cdot \mDelta_{i}.
$$
\fi
Note that even when $\mDelta_i^\gamma=\textbf{0}$ for any $i\in[n]$, the $i$-th record still influences the BiMax-M objective and needs to be taken into account when solving the outer optimization in BiMax-M.

\textbf{Finding \texttt{MaxS-M}}. We first find $\texttt{MaxS-M}(\gamma)$ by solving the following optimization problem: 
    \begin{equation*}
\begin{array}{ll@{}ll}
\underset{\mDelta\in\mathcal{R}^{n\times d}}{\arg\max}  & \sum_{i\in[n]}\mG_i\cdot\mDelta_i\\[10pt]
\textnormal{s. t.}& \normx{\mDelta_i}\leq \gamma, &\forall i\in[n].
\end{array}
\end{equation*}
In this problem, the constraints are independent for each $i\in[n]$, $\normx{\mG_i}\neq 0$, and the objective is simply a summation across $\mG_i\cdot(\mD_i+\mDelta_i)$ values, so that a solution $\mDelta^*$ is optimal for this problem if and only if for each $i\in[n]$,  $\mDelta_i^*$ is an optimal solution to
    \begin{equation*}
\begin{array}{ll@{}ll}
\underset{\mDelta\in\mathcal{R}^{d}}{\arg\max}  &\mG_i\cdot\mDelta_i\\[10pt]
\textnormal{s. t.}& \normx{\mDelta_i}\leq \gamma.
\end{array}
\end{equation*}

Solving this problem, we have: 

\begin{lemma}\label{lemma:KKT_magnitude}
    For any $i\in[n]$, whenever $\normx{\mG_i}\neq 0$ and $\gamma\geq 0$, the optimal solution to
    \begin{equation*}
\begin{array}{ll@{}ll}
\underset{\mDelta\in\mathcal{R}^{d}}{\arg\min}  & -\mG_i\cdot\mDelta_i\\[10pt]
\textnormal{s. t.}& \normx{\mDelta_i}\leq \gamma,
\end{array}
\end{equation*}
is
\[
\mDelta_i^\gamma=\gamma\frac{\mG_i}{\normx{\mG_i}}.
\]
\end{lemma}

\textbf{Solving BiMax-M}. The goal is now to find a parameter $\gamma\geq0$ such that the maximum number of validation queries are answered correctly. Observe that, by definition, the $i$-th validation query, $i\in[n_V]$ is correctly answered when $\forall j\in[n_V]\setminus Y_i^V$,
%$$(\vq_i\cdot(\mD_{Y_i}+\gamma \mG_{Y_i}-\mD_{j}-\gamma \mG_{j})) \geq 0,$$
%Or equivalently 
\if 0
$$\alpha (\mQ_i^V\cdot \mG_{g}- \mQ_i^V\cdot \mG_{j}) > \mQ_i^V\cdot \mD_{j}-\mQ_i^V\cdot \mD_{j^*} \quad\quad \forall j\in[n_V]\setminus j^*,$$
where $j^*=Y_i^V$. Equivalently, $i$-th query is answered correctly when $\forall j\in[n_V]\setminus j^*$
\fi
\begin{align*}
    \mQ_i^V\cdot \mD_{Y_i}+\gamma \mathds{I}[\normx{\mG_{Y_i^V}}\neq0]\mQ_i^V\cdot \frac{\mG_{Y_i^V}}{\normx{\mG_{Y_i^V}}} > \mQ_i^V\cdot \mD_{j}+\gamma\mathds{I}[\normx{\mG_{j}}\neq0]\mQ_i^V\cdot \frac{\mG_{j}}{\normx{\mG_{j}}}.
\end{align*}
where, we abuse notation and assume $\frac{\mathds{I}[\normx{\mG_{j}}\neq0]}{\normx{\mG_{j}}}=0$ if $\normx{\mG_{j}}=0$ for any $j$. Define $$\mathcal{G}_{i, j}=\mathds{I}[\normx{\mG_{j}}\neq0]\mQ_i^V\cdot \frac{\mG_{j}}{\normx{\mG_{j}}}, \quad\text{and}\quad \mathcal{S}_{i, j}=\mQ_i^V\cdot \mD_{j}$$ for any $j\in[n]$. Thus, we have that the $i$-th validation query is answered correctly when $\forall j\in[n_V]\setminus Y_i^V$
\[
\gamma\in I_{i, j}=\begin{cases}
    \Bigl(-\infty, \frac{\mathcal{S}_{i, j}-\mathcal{S}_{i, Y_i^V}}{\QG{}}\Bigr),& \text{if } \QG{}<0\\[5pt]
    \Big(\frac{\mathcal{S}_{i, j}-\mathcal{S}_{i, Y_i^V}}{\QG{}}, \infty\Big),              & \text{if } \QG{}>0\\[5pt]
    \mathcal{R},              & \text{if } \QG{}=0 \text{ and } \mathcal{S}_{i, j}-\mathcal{S}_{i, Y_i^V}< 0 \\
    \emptyset,              & \text{otherwise.}
\end{cases}
\]
\if 0
\[
\alpha\in I_{i, j}=\begin{cases}
    \Bigl(-\infty, \frac{\mQ_i^V\cdot \mD_{j}-\mQ_i^V\cdot \mD_{j^*}}{\mQ_i^V\cdot \mG_{j^*}- \mQ_i^V\cdot \mG_{j}}\Bigr),& \text{if } \mQ_i^V\cdot \mG_{j^*}- \mQ_i^V\cdot \mG_{j}<0\\[5pt]
    \Big(\frac{\mQ_i^V\cdot \mD_{j}-\mQ_i^V\cdot \mD_{j^*}}{\mQ_i^V\cdot \mG_{j^*}- \mQ_i^V\cdot \mG_{j}}, \infty\Big),              & \text{if } \mQ_i^V\cdot \mG_{j^*}- \mQ_i^V\cdot \mG_{j}>0\\[5pt]
    \mathcal{R},              & \text{if } \mQ_i^V\cdot \mG_{j^*}- \mQ_i^V\cdot \mG_{j}=0 \text{ and } \mQ_i^V\cdot \mD_{j}-\mQ_i^V\cdot \mD_{j^*}< 0 \\
    \emptyset,              & \text{otherwise.}
\end{cases}
\]
\fi
$I_{i, j}$ defines an interval in $\mathcal{R}$ so that $\gamma\in I_{i, j}$ means the correct answer to the $i$-th validation query has higher similarity to the query compared with the $j$-th data record, for $j\neq Y_i^V$. Thus, to answer the $i$-th  query correctly, we must have $\gamma\in I_i=\cap_{j\neq Y_i^V}I_{i, j}$. %In other words, the $i$-th query is answered correctly whenever $\gamma\in I_i$ where $I_i$ is an interval in $\mathcal{R}$. 
To maximize the number of queries answered correctly, after finding $I_i$ for all $i$, we simply need to find a $\gamma$ value that intersects the most number of intervals among $I_1, ..., I_{n_V}$, i.e., $\arg\max_{\gamma}\sum_{I_i\in I}\mathds{I}[\gamma\in I_i]$, where $I=\{I_1, ..., I_{n_V}\}$. Finding a point where maximum intervals overlap is a basic algorithmic problem and can be done by a single iteration through the intervals after sorting their start and end points. 

\textbf{\name{}-M}. Alg.~\ref{alg:optimval_val} presents \name{}-M, which formalizes the above procedure, and also incorporates the constraint $\gamma\geq 0$. Lines~\ref{alg:line:for_qv}-\ref{alg:line:add_R} find the intervals $I_i$, $\forall i\in[n_V]$ and add it to a list $I$ (intersection of half intervals can be found by calculating the maximum of lower bounds and minimum of upper bounds of the intervals, done in lines~\ref{alg:line:p}-~\ref{alg:line:bounds}). The algorithm returns new data embeddings by simply performing a single addition.

\textbf{Time complexity}. Alg.~\ref{alg:optimval_val} can be implemented so that lines \ref{alg:line:l_u}-\ref{alg:line:check_neg} with a single pass over the dataset. Therefore, finding $I_i$ for all queries (lines~\ref{alg:line:for_qv}-\ref{alg:line:add_R}) take $O(n_V\times n\times d)$. Calculating $G$ takes time $O(n_T\times d)$. Finding $\gamma^*$ in line~\ref{alg:line:final_gamma} is the basic problem of finding the maximum number of overlapping ranges and can be done in $O(n_V\log(n_V))$, by first sorting the ranges in $I$ based on their lower bound and iteratively traversing the sorted list and keeping track of the number of overlapping ranges. Thus, Alg.~\ref{alg:optimval_val} can be implemented in $O(n_V(n\times d+ \log(n_V)+n_T\times d)$.

\subsection{Proof of Theorem~\ref{thm:nopet_n}}\label{proof:them:nopetn}
\textbf{Setup}. In the discussion below, for simplicity, we assume for any $i\in[n]$ $\mG_i\cdot \mD_i\geq 0$. Although the discussion below can be extended to consider $\mG_i\cdot \mD_i< 0$, the results will be more tedious (see Lemma~\ref{lemma:KKT_norm}). Moreover, in practice we expect $\mG_i\cdot \mD_i\geq 0$ to hold, as otherwise queries and their correct answers will have negative similarity suggesting either a poor training dataset or pre-trained embedding model. We simply set $\mG_i=\textbf{0}$ whenever $\mG_i\cdot \mD_i< 0$.
Moreover, similar to Appx.~\ref{proof:them:nopetm}, we rewrite the \texttt{MaxS-N} objective as
$$
\underset{\mDelta\in\mathcal{R}^{n\times d}}{\arg\max} \sum_{i\in[n_T]}\mQ^T_i\cdot (\mD_{Y_i^T}+\mDelta_{Y_i^T})=\underset{\mDelta\in\mathcal{R}^{n\times d}}{\arg\max} \sum_{i\in[n]}\mG_i\cdot \mDelta_{i}=\underset{\mDelta\in\mathcal{R}^{n\times d}}{\arg\max} \sum_{i\in[n], \normx{\mG_i}\neq0}\mG_i\cdot \mDelta_{i},
$$
and set $\mDelta_i^\gamma=\textbf{0}$ for any $i$ where $\normx{\mG_i}=0.$ Furthermore, we assume the embeddings $\mD_i$ are normalized, (i.e., $\normx{\mD_i}=1$, $\forall i\in [n]$), or we otherwise normalize them.

\textbf{Finding \texttt{MaxS-N}}. We first find $\texttt{MaxS-N}(\gamma)$ by solving the following optimization problem: 
    \begin{equation*}
\begin{array}{ll@{}ll}
\underset{\mDelta\in\mathcal{R}^{n\times d}}{\arg\max}  & \sum_{i\in[n]}\mG_i\cdot\mD_i\\[10pt]
\textnormal{s. t.}& \normx{\mDelta_i}^2\leq \gamma, &\forall i\in[n]\\
                 & \normx{\mDelta_i+\mD_i}= 1, &\forall i\in[n].
\end{array}
\end{equation*}
In this problem, the constraints are independent for each $i\in[n]$, and the objective is simply a summation across $\mG_i\cdot(\mD_i+\mDelta_i)$ values, so that a solution $\mDelta^*$ is optimal for this problem if and only if for each $i\in[n]$,  $\mDelta_i^*$ is an optimal solution to
    \begin{equation*}
\begin{array}{ll@{}ll}
\underset{\mDelta\in\mathcal{R}^{d}}{\arg\max}  &\mG_i\cdot\mD_i\\[10pt]
\textnormal{s. t.}& \normx{\mDelta_i}^2\leq \gamma,\\
                 & \normx{\mDelta_i+\mD_i}= 1.
\end{array}
\end{equation*}

Solving this problem, we have: 
\begin{lemma}\label{lemma:KKT_norm}
    For any $i\in[n]$, let $\theta_i$ be the angle between $\mG_i$ and $\mD_i$. Whenever $\theta_i\in[0, \frac{\pi}{2}]$, $\gamma\geq0$, $\normx{\mD_i}=1$ and $\normx{\mG_i}\neq 0$ the optimal solution to the optimization problem
    \begin{equation*}
\begin{array}{ll@{}ll}
\underset{\mDelta\in\mathcal{R}^{d}}{\arg\min}  & -\mG_i\cdot\mDelta_i\\[10pt]
\textnormal{s. t.}& \normx{\mDelta_i}^2\leq \gamma,\\
                 & \normx{\mDelta_i+\mD_i}= 1,
\end{array}
\end{equation*}
is
\[
\mDelta_i^\gamma=\begin{cases*}
    \frac{\mG_i}{\normx{\mG_i}}-\mD_i,              & $\textnormal{if}\; \cos\theta_i\geq 1-\frac{\gamma}{2}$ \\
    \frac{\sqrt{\gamma(4-\gamma)}(\mG_i-(\mD_i\cdot\mG_i)\mD_i)}{2\normx{\mG_i-(\mD_i\cdot\mG_i)\mD_i}}-\frac{\gamma}{2}\mD_i,              & \textnormal{otherwise.}
\end{cases*}
\]
\end{lemma}
For simplicity of notation, we denote by $\mZ_i\in \mathcal{R}^d$ the vector 
$$
\mZ_i=\frac{\mG_i-(\mD_i\cdot\mG_i)\mD_i}{\normx{\mG_i-(\mD_i\cdot\mG_i)\mD_i}}.
$$
We also denote by $\mathcal{\mZ}_{i, j}=\mQ_i^V\cdot\mZ_j$, by $\mathcal{\mD}_{i, j}=\mQ_i^V\cdot\mD_j$ and by $\mathcal{\mG}_{i, j}=\mathds{I}[\normx{\mG_j}\neq 0]\mQ_i^V\cdot\frac{\mG_j}{\normx{\mG_j}}$ for all $i\in[n_V], j\in[n]$.

\textbf{Solving BiMax-N}. Next, we consider solving BiMax-N, $\forall j\in[n]$. Substituting $\Delta^{\gamma}$ into the definition of $\mathcal{I}_i^V(\mDelta^\gamma)$ for each $i\in [n_V]$, $\mathcal{I}_i^V(\mDelta^\gamma)=1$ if and only if for all $j\in[n]\setminus Y_i^V$
\begin{align}\label{eq:I:basic_with_gamma}
    \mQ^V_i\cdot(\mD_{Y_i^V}+\mDelta_{Y_i^V}^\gamma) > \mQ^V_i\cdot(\mD_{j}+\mDelta_{j}^\gamma).
\end{align}
First, consider the case simpler setting when $\normx{\mG_j}\neq 0$ for all $j\in[n]$. Let $\bar{\theta}_{i, j}=\max\{\theta_{i}, \theta_{j}\}$ and $\ubar{\theta}_{i, j}=\min\{\theta_{i}, \theta_{j}\}$. Substituting the values from Lemma~\ref{lemma:KKT_norm}, we have Eq.~\ref{eq:I:basic_with_gamma} is equivalent to
\begin{equation}\label{eq:I:gamma_cases}
%\hspace{-0.5cm}
\begin{cases*}
    (1-\frac{\gamma}{2})\mathcal{\mD}_{i, Y_i^V}+\frac{\sqrt{\gamma(4-\gamma)}}{2}\mathcal{\mZ}_{i, Y_i^V} > (1-\frac{\gamma}{2})\mathcal{\mD}_{i, j}+\frac{\sqrt{\gamma(4-\gamma)}}{2}\mathcal{\mZ}_{i, j},              & 
    $\text{if}\; \gamma<2-2 \cos\ubar{\theta}_{Y_i^V, j}$\\%[8pt]
    (1-\frac{\gamma}{2})\mathcal{\mD}_{i, Y_i^V}+\frac{\sqrt{\gamma(4-\gamma)}}{2}\mathcal{\mZ}_{i, Y_i^V} > \mathcal{\mG}_{i, j},              & \parbox[t]{4.3cm}{if $\gamma\geq2-2 \cos\ubar{\theta}_{Y_i^V, j}$, \hspace{2cm}$\gamma<2-2 \cos\bar{\theta}_{Y_i^V, j}$, $\theta_{Y_i^V}<\theta_j$}\\%[8pt]
    \mathcal{\mG}_{i, Y_i^V} > (1-\frac{\gamma}{2})\mathcal{\mD}_{i, j}+\frac{\sqrt{\gamma(4-\gamma)}}{2}\mathcal{\mZ}_{i, j},              & \parbox[t]{4.3cm}{if $\gamma\geq2-2 \cos\ubar{\theta}_{Y_i^V, j}$, \hspace{2cm}$\gamma<2-2 \cos\bar{\theta}_{Y_i^V, j}$, $\theta_{Y_i^V}>\theta_j$}\\%[8pt]
    \mathcal{\mG}_{i, Y_i^V} > \mathcal{\mG}_{i, j},              & \text{otherwise.} 
\end{cases*}
\end{equation}

Observe that finding the values of $\gamma$ for which the inequalities in Eq.~\ref{eq:I:gamma_cases} hold requires solving inequalities of the form 
$a\sqrt{\gamma(4-\gamma)}+b\gamma+c>0, 
$ for some $a,b,c\in\mathcal{R}$.
\begin{lemma}\label{lemma:quadratic_sol}
    Let $f(\gamma)=a\sqrt{\gamma(4-\gamma)}+b\gamma+c$ for $a,b,c\in\mathcal{R}$. We say there are $k$ roots, $\gamma_0, ..., \gamma_{k-1}$, if there exists $k$ distinct $\gamma_i\in(0, 4)$ for which $f(\gamma_i)=0$. We must have $k\in\{0, 1, 2\}$, if two roots exist, we have$$\gamma_0=\frac{2a^2-bc-a\sqrt{\delta}}{a^2+b^2}, \; \gamma_1=\frac{2a^2-bc+a\sqrt{\delta}}{a^2+b^2},\;\textnormal{for } \delta=4a^2-4bc-a^2c^2,$$ and if only one root exists 
    $$\gamma_0\in\{\frac{2a^2-bc-a\sqrt{\delta}}{a^2+b^2},\frac{2a^2-bc+a\sqrt{\delta}}{a^2+b^2}\}.$$
    Moreover, for any $\gamma\in(0,4)$, $f(\gamma)>0$ holds if and only if
\begin{equation}\label{eq:gamm_ineq_cases}\hspace{-0.5cm}
\gamma\in\begin{cases*}
 (0, 4)& \textnormal{if zero roots and } $(c>0$ \textnormal{or} $(c=0, b>0)$ \textnormal{or} $(c=b=0, a>0))$\\
 \mathcal{\emptyset} &\textnormal{if zero roots and } $(c<0$ \textnormal{or} $(c=0, b<0)$ \textnormal{or} $(c=b=0, a\leq0))$\\
 (\gamma_0, \gamma_1) &\textnormal{if two roots and} $c<0$\\
 (0, \gamma_0)\cup (\gamma_1, 4) &\textnormal{if two roots and} $c>0$\\
 (0, \gamma_0) &\textnormal{if one root and} $(c>0$ \textnormal{or} $c=0,a>0)$\\
  (\gamma_0,4) &\textnormal{if one root and} $(c<0$ \textnormal{or} $c=0,a<0).$\\
\end{cases*}
\end{equation}
%\textsc{SolveGamma}($a$, $b$, $c$) refers to an algorithm that uses Eq.~\ref{eq:gamm_ineq_cases} to solve $a\sqrt{\gamma(4-\gamma)}+b\gamma+c>0$.
\end{lemma}

Thus, for the $i$-th query and the $j$-th data record, $i\in[n_V]$, $j\in[n]\setminus Y_i^V$, we can use Eq.~\ref{eq:gamm_ineq_cases} together with Eq.~\ref{eq:I:gamma_cases} to obtain a set $I_{i, j}\subseteq{(0, 4)}$ such that Eq.~\ref{eq:I:basic_with_gamma} holds if and only if $\gamma\in 
I_{i, j}$. Moreover, $I_{i, j}$ will consist of at most 3 intervals in $\subseteq{(0, 4)}$. Let $I_i=\cap_{j\in[n]\neq Y_i^V} I_{i, j}$, and note that the $i$-th query will be answered accurately if and only if $\gamma\in I_i$. Finally, $\gamma^*=\arg\max_{\gamma}\sum_{i\in[n_V]}\mathds{I}[\gamma\in I_i]$ is an optimal solution to the MAxSS problem. 

Now whenever $\normx{\mG_j}=0$ for any $j\in[n]$, Eq.~\ref{eq:I:basic_with_gamma} changes since we need to also consider $\mDelta_{Y_i^V}^\gamma=0$, $\mDelta_{j}^\gamma=0$ or both. Each case leads to a similar set of inequalities to Eq.~\ref{eq:I:gamma_cases}, which are similarly solved using Lemm~\ref{lemma:quadratic_sol}. Moreover, we calculate validation accuracy separately for when $\gamma=0$. 
%This procedure is shown in Alg XXXXXXXXX.

\textbf{\name{}-N}. \name{}-N follows the above procedure. It first calculates $\mG$ and $\mZ$, so that for the $i$-th query, and $j$-th data points, $j\neq Y_i^V$ applying Lemma~\ref{lemma:quadratic_sol} to Eq.~\ref{eq:I:gamma_cases} gives $I_{i, j}$. Then, it computes $I_{i}$ by a single pass over $I_{i, j}$ values, and finally finds a $\gamma $ value that intersects most $I_i$'s by sorting their beginning and end and iteration through the list. 

\textbf{Time Complexity}. Calculating $\mG$ takes time $O(dn_T)$, and $\mZ$ takes $O(dn)$. Then, for the $i$-th query, it finds $I_i$ which takes $O(nd)$, since each $I_{i, j}$ only contains a constant number of intervals and can be computed in $O(d)$. Finally, finding $\gamma^*$ that intersects the most intervals, can be done in $O(n_V\log n_V)$. Thus, in total, \name{}-N takes total of $O(n_V(nd+\log n_V)+n_Td)$.

\subsection{Proof of Technical Lemmas}\label{appx:lemmas}
\subsubsection{Proof of Lemma~\ref{lemma:KKT_magnitude}}
If $\gamma=0$ the only feasible, and therefore optimal, solution is $\mDelta=\textbf{0}$. Next, consider $\gamma>0$.  Observe that the optimization problem is convex, so we find $\mDelta_i$ values that satisfy the KKT conditions which, since the Slater's condition holds whenever $\gamma>0$, provide necessary and sufficient conditions for optimality. %The problem is equivalent to minimizing $$-\sum_{i\in [n]}\mG_i\cdot\mD_i.$$ 

We have that the Lagrangian is
$$
L(\mDelta_i, \mu, \lambda)=-\mG_i\cdot\mD_i+\frac{1}{2}\mu (\normx{\mDelta_i}^2-\gamma^2),
$$
so that the KKT conditions are 
\begin{align}
-\mG_i+\mu\mDelta_i&=\textbf{0},\label{eq:kkt_m_lagrangian_grad}\\
    \mu (\normx{\mDelta_i}^2-\gamma^2)&=0,\label{eq:m:norm_slack}\\
    \mu\geq 0 \label{eq:m:dual_feas}.
\end{align}
If $\mu=0$, we must have $\normx{\mG_i}=0$, which is a contradiction. Thus, $\mu>0$, and  
\begin{align}\label{eq:m:contraint_w_equality}
\normx{\mDelta_i}^2-\gamma^2=0. 
\end{align}
From Eq.~\ref{eq:kkt_m_lagrangian_grad}, 
\begin{align}\label{eq:m:delta_w_mu}
    \mDelta_i=\frac{\mG_i}{\mu}
\end{align}
Substituting this into Eq.~\ref{eq:m:contraint_w_equality}, and since $\mu>0$, we have
$$
\mu=\frac{\normx{\mG_i}}{\gamma},
$$
and substituting back into Eq.~\ref{eq:m:delta_w_mu}, we have
$$
\mDelta_i=\gamma\frac{\mG_i}{\normx{\mG_i}}.
$$

\subsubsection{Proof of Lemma~\ref{lemma:KKT_norm}}
We replace $\normx{\mDelta_i+\mD_i}= 1$ with $\normx{\mDelta_i+\mD_i}\leq 1$ so that the optimization is over a convex set, and solve the problem with $\normx{\mDelta_i+\mD_i}\leq 1$ constraint. As we will see, the only solution to the problem will have $\normx{\mDelta_i+\mD_i}= 1$, implying that it is the optimal solution to the original problem as well. Thus, consider the convex problem with $\normx{\mDelta_i+\mD_i}\leq 1$ constraint, or equivalantly $\normx{\mDelta_i+\mD_i}^2\leq 1$. 

We find $\mDelta_i$ values that satisfy the KKT conditions, which, since the Slater's condition holds whenever $\gamma>0$, provide necessary and sufficient conditions for optimality. %The problem is equivalent to minimizing $$-\sum_{i\in [n]}\mG_i\cdot\mD_i.$$ 

We have that the Lagrangian is
$$
L(\mDelta_i, \mu, \lambda)=-\mG_i\cdot\mD_i+\frac{1}{2}\mu (\normx{\mDelta_i}^2-\gamma)+\frac{1}{2}\lambda(\normx{\mDelta_i}^2+2\mDelta_i\cdot\mD_i),
$$
so that the KKT conditions are 
\begin{align}
    -\mG_i+\mu\mDelta_i+\lambda\mDelta_i+\lambda\mD_i&=\textbf{0},\label{eq:kkt_norm_lagrangian_grad}\\
    \lambda(\normx{\mDelta_i}^2+2\mDelta_i\cdot\mD_i)&=0,\label{eq:norm:bound_slack}\\
    \mu (\normx{\mDelta_i}^2-\gamma)&=0,\label{eq:norm:norm_slack}\\
    \normx{\mDelta_i}^2&\leq \gamma,\label{eq:norm:primal_feas_norm}\\
    \normx{\mDelta_i}^2+2\mDelta_i\cdot\mD_i&\leq 0,\label{eq:norm:primal_feas}\\
    \mu,\lambda&\geq 0 \label{eq:norm:dual_feas},
\end{align}
where Eq.~\ref{eq:kkt_norm_lagrangian_grad} is obtained by setting the gradient of the Lagrangian to zero, and Eq.~\ref{eq:norm:primal_feas} and ~\ref{eq:norm:bound_slack} are obtained by substituting $$\normx{\mD_i+\mDelta_i}^2=\normx{\mD_i}^2+\normx{\mDelta_i}^2+2\mDelta_i\cdot\mD_i=1+\normx{\mDelta_i}^2+2\mDelta_i\cdot\mD_i,$$
since $\normx{\mDelta_i}=1$.

To find the points satisfying all Eq.~\ref{eq:kkt_norm_lagrangian_grad}-\ref{eq:norm:dual_feas}, we consider 4 setting depending on whether $\lambda=0$ or $\mu=0$ or not.

\textbf{Case 1, $\lambda=0, \mu=0$}. Substituting $\lambda=0$ and $\mu=0$ in Eq.~\ref{eq:kkt_norm_lagrangian_grad}, we have $\mG_i=\mathbf{0}$. Thus, no solution with $\lambda=0$ and $\mu=0$ exists since $\mG_i\neq\mathbf{0}$. 

\textbf{Case 2, $\lambda=0, \mu>0$}. Having $\lambda=0$ and $\mu>0$, we have from Eq.~\ref{eq:kkt_norm_lagrangian_grad}
$$
\mDelta_i=\frac{1}{\mu}\mG_i,
$$
substituting which into Eq.~\ref{eq:norm:norm_slack} with $\mu>0$ w have
\begin{align*}
\frac{1}{\mu^2}\normx{\mG_i}^2=\gamma,\\
\mu=\frac{\normx{\mG_i}}{\sqrt{\gamma}},
\end{align*}
and therefore $\mDelta_i=\sqrt{\gamma}\frac{\mG_i}{\normx{\mG_i}}$ (KKT conditions become similar to the ones in Theorem~\ref{thm:nopet}). Because $\mu>0$, we must have $\normx{\mDelta_i}^2=\gamma$ due to Eq.~\ref{eq:norm:norm_slack}, so that from Eq.~\ref{eq:norm:primal_feas} we have 
\begin{align*}
\gamma+2\sqrt{\gamma}\frac{\mG_i}{\normx{\mG_i}}\cdot\mD_i\leq 0,   \\ 
\gamma+2\sqrt{\gamma}\cos\theta_i\leq 0,    
\end{align*}
where $\theta_i$ is the angle between $\mG_i$ and $\mD_i$. Thus, we must have
$$
\cos \theta_i\leq -\frac{\sqrt{\gamma}}{2}.
$$
Since $\cos \theta_i\geq0$ and $\gamma>0$ by assumption, there are no solutions with  $\lambda=0$, $\mu>0$.

\textbf{Case 3, $\lambda>0, \mu=0$}. Substituting $\mu=0$ in Eq.~\ref{eq:kkt_norm_lagrangian_grad} we have
$$
\mDelta_i=\frac{\mG_i-\lambda\mD_i}{\lambda}.
$$
Substituting this in Eq.~\ref{eq:norm:bound_slack}, we have
\begin{align*}
\normx{\frac{\mG_i-\lambda\mD_i}{\lambda}}^2+2\frac{\mG_i-\lambda\mD_i}{\lambda}\cdot\mD_i=0,\\
\normx{\mG_i}^2+\lambda^2-2\lambda\mG_i\cdot\mD_i+2\lambda(\mG_i\cdot\mD_i-\lambda)=0,
\end{align*}
implying, since $\lambda>0$,
$$
\lambda=\normx{\mG_i},
$$
and therefore
$$
\mDelta_i=\frac{\mG_i}{\normx{\mG_i}}-\mD_i.
$$
Note that to satisfy Eq.~\ref{eq:norm:primal_feas_norm}, we must have
\begin{align*}
(\frac{\mG_i}{\normx{\mG_i}}-\mD_i)\cdot(\frac{\mG_i}{\normx{\mG_i}}-\mD_i)\leq \gamma,
\end{align*}
simplifying which we have
$$
\cos\theta_i\geq 1-\frac{\gamma}{2}.
$$
To summarize, we showed that setting
$$
\mDelta_i=\frac{\mG_i}{\normx{\mG_i}}-\mD_i,\; \mu=0,\; \text{and}\; \lambda=\normx{\mG_i}
$$
satisfy all KKT conditions whenever $\cos\theta_i\geq 1-\frac{\gamma}{2}$.

\textbf{Case 4, $\lambda>0, \mu>0$}. From Eq.~\ref{eq:kkt_norm_lagrangian_grad}, we have
\begin{align}
\mDelta_i=\frac{\mG_i-\lambda\mD_i}{\lambda+\mu}    \label{eq:delta_case4_with_mu_lambda}
\end{align}
and since $\normx{\mDelta_i}-\gamma=0$, Eq.~\ref{eq:norm:bound_slack} simplifies to $\gamma+2\mDelta_i\cdot\mD_i=0$, so that
$$
2\frac{\mG_i\cdot\mD_i-\lambda\mD_i\cdot\mD_i}{\lambda+\mu}=-\gamma.
$$
Therefore,
\begin{align}\label{eq:case4_mu}
\mu=2\frac{\lambda-\mG_i\cdot\mD_i}{\gamma}-\lambda.    
\end{align}
Substituting this in Eq.`\ref{eq:delta_case4_with_mu_lambda}, we have
\begin{align}\label{eq:delta:case4_with_lambda}
\mDelta_i=\frac{\gamma}{2}\frac{\mG_i-\lambda\mD_i}{\lambda-\mG_i\cdot\mD_i}.
\end{align}
Now, substituting Eq.~\ref{eq:delta:case4_with_lambda} in Eq.~\ref{eq:norm:norm_slack}, we get 
$$
\frac{\gamma^2}{4(\lambda-\mG_i\cdot\mD_i)^2}(\mG_i-\lambda\mD_i)\cdot(\mG_i-\lambda\mD_i)=\gamma
$$
So
$$
\frac{\gamma}{4}(\normx{\mG_i}^2-\lambda^2-2\lambda\mG_i\cdot\mD_i)=\lambda^2+(\mG_i\cdot\mD_i)^2-2\lambda\mG_i\cdot\mD_i.
$$
Rearranging, we obtain a quadratic equation in $\lambda$,
$$
\lambda^2-2\mD_i\cdot\mG_i\lambda+\frac{\normx{\mG_i}^2(\gamma-4\cos^2\theta_i)}{\gamma-4}=0,
$$
where we have used the fact that $(\mD_i\cdot\mG_i)^2=\normx{\mG_i}^2\cos^2\theta_i$. Solving the equation and simplifying we have
\begin{align}\label{eq:case4_lambda_root}
    \lambda=\mD_i\cdot\mG_i\pm\normx{\mG_i}\sin\theta_i\sqrt{\frac{\gamma}{4-\gamma}}.
\end{align}
Note that we must also have $\mu>0$, so substituting Eq.~\ref{eq:case4_lambda_root} into Eq.~\ref{eq:case4_mu} and rearranging we must have 
\begin{align}
\pm\normx{\mG_i}\sin\theta_i\sqrt{\frac{\gamma}{4-\gamma}}(\frac{2}{\gamma}-1)-\mD_i\cdot\mG_i>& \,0,\notag\\
\pm\sin\theta_i\sqrt{\frac{\gamma}{4-\gamma}}(\frac{2}{\gamma}-1)>&\cos\theta_i.\label{eq:case_4:mu_positive}
\end{align}
Observe that, for $\theta_i\in[0, \frac{\pi}{2}]$, both $\sin\theta_i$ and $\cos\theta_i$ are non-negative, so that when $\gamma<2$ only the positive branch is able to satisfy Eq.~\ref{eq:case_4:mu_positive}. In this case, we have 
\begin{align}\label{eq:case4_lambda_correct_root}
    \lambda=\mD_i\cdot\mG_i+\normx{\mG_i}\sin\theta_i\sqrt{\frac{\gamma}{4-\gamma}}.
\end{align}
Note that 
\begin{align*}
    \normx{\mG_i}\cos\theta_i+\normx{\mG_i}\sin\theta_i\sqrt{\frac{\gamma}{4-\gamma}}>0,
\end{align*}
for $\theta_i\in[0, \frac{\pi}{2}]$, and thus $\lambda>0$ is satisfied. Simplifying Eq.~\ref{eq:case_4:mu_positive}, observe that for $\theta_i\in[0, \frac{\pi}{2}]$ and $0<\gamma< 2$,
\begin{align}\label{eq:norm:trig_simp}
\sin\theta_i\sqrt{\frac{(2-\gamma)^2}{\gamma(4-\gamma)}}>\cos\theta_i\iff\theta_i>\arctan(\sqrt{\frac{\gamma(4-\gamma)}{(2-\gamma)^2}})\iff\cos\theta_i<\frac{1}{\sqrt{1+\frac{\gamma(4-\gamma)}{(2-\gamma)^2}}}=1-\frac{\gamma}{2}.
\end{align}
To summarize, taking the positive branch in Eq.~\ref{eq:case4_lambda_root}, we showed that 
\begin{align*}
    \mDelta_i&=\frac{\sqrt{\gamma(4-\gamma)}(\mG_i-(\mD_i\cdot\mG_i)\mD_i)}{2\normx{\mG_i}\sin\theta_i}-\frac{\gamma}{2}\mD_i,\\ \lambda&=\mD_i\cdot\mG_i+\normx{\mG_i}\sin\theta_i\sqrt{\frac{\gamma}{4-\gamma}},\text{and}\\
    \mu&=\normx{\mG_i}\sin\theta_i\sqrt{\frac{(2-\gamma)^2}{\gamma(4-\gamma)}}-\mD_i\cdot\mG_i
\end{align*}
satisfy the KKT conditions whenever 
$$
\cos\theta_i<1-\frac{\gamma}{2}.
$$
Finally, consider the negative branch in Eq.~\ref{eq:case4_lambda_root} and $\gamma>2$. To have $\mu>0$, following an argument similar to Eq.~\ref{eq:norm:trig_simp}, we must have
\begin{align*}
    \cos\theta_i<|1-\frac{\gamma}{2}|, 
\end{align*}
and similarly for $\lambda>0$, we must have
\begin{align*}
    \normx{\mG_i}\cos\theta_i-\normx{\mG_i}\sin\theta_i\sqrt{\frac{\gamma}{4-\gamma}}>0\iff \theta_i<\arctan(\sqrt{\frac{4-\gamma}{\gamma}})\iff \cos\theta_i>\frac{1}{2}\sqrt{\gamma}.
\end{align*}
However, $\frac{\sqrt{\gamma}}{2}\geq|1-\frac{\gamma}{2}|$ for all $\gamma\in[2, 4]$, which implies $\mu>0$ and $\lambda>0$ cannot hold at the same time in this case. Finally, observe that $\normx{\mG_i-(\mD_i\cdot\mG_i)\mD_i}=\normx{\mG_i}\sin \theta_i$.
\qed

\subsubsection{Proof of Lemma~\ref{lemma:quadratic_sol}}
We first find $\gamma$ values where $f(\gamma)=0$. Observe that 
$$
a\sqrt{\gamma(4-\gamma)}+b\gamma+c=0\Rightarrow a^2\gamma(4-\gamma)-(b\gamma+c)^2=0,
$$
so that $a^2\gamma(4-\gamma)-(b\gamma+c)^2=0$ is necessary for $f(\gamma)=0$. Moreover solving the quadratic inequality, $a^2\gamma(4-\gamma)-(b\gamma+c)^2=0$ iff $\gamma\in\{\gamma_0,\gamma_1\}$ where
$$\gamma_0=\frac{2a^2-bc-a\sqrt{\delta}}{a^2+b^2}, \; \gamma_1=\frac{2a^2-bc+a\sqrt{\delta}}{a^2+b^2},\;\textnormal{for } \delta=4a^2-4bc-a^2c^2.$$
Thus, $\gamma\in\{\gamma_0,\gamma_1\}$ is necessary for $f(\gamma)=0$, which proves the first part of the lemma.

The proof of the rest of the lemma uses the above, the fact that $f$ is continuous, in addition to the following facts:  
\begin{align}
    &c>0 \Rightarrow f(0)>0, \;c<0 \Rightarrow f(0)<0\label{eq:check_c}\\
    &c=0, b>0 \Rightarrow f(4)>0, \;c=0, b<0 \Rightarrow f(4)<0\label{eq:c0_checkb}\\
    &f \text{ has at most 1 local extrema, which is a maximum if } a>0 \text{ and a minimum if } a<0\label{eq:check_a}
\end{align}

If there are zero roots, $f(\gamma)>0$ either for all of $(0, 4)$ or none of it, which can be determined by checking the function value at 0 or 4 using Eq.~\ref{eq:check_c} and \ref{eq:c0_checkb} if $c\neq 0$ or $b\neq 0$. If $c=b=0$, then we can solve $f(\gamma)>0$ based on whether the function has a minimum or maximum using Eq.~\ref{eq:check_a}, or is a constant when $a=0$.

If there is one root, $\gamma_0$, then $f(\gamma)$ is either positive after $\gamma_0$ or before it. If $c\neq 0$, we can find this by checking $f(0)$ using Eq.~\ref{eq:check_c}. If $c=0$, then $f(\gamma)$ has two roots, at $0$ and $\gamma_0$, and $f(\gamma)>0$ can be determined based on whether the function has a minimum or maximum using Eq.~\ref{eq:check_a}. Both $c$ and $a$ cannot be zero, because the function $b\gamma$ cannot be zero both at $\gamma=0$ and $\gamma=\gamma_0$ for $\gamma_0\neq 0$.

If there are two roots, then $f(\gamma)>0$ either between the two roots, or outside the interval between the two roots, which can be checked based on $f(0)$ using Eq~\ref{eq:check_c}. $c$ cannot be zero because $f(\gamma)$ cannot be zero at $\gamma=0$, $\gamma=\gamma_0$ and $\gamma=\gamma_1$ for 3 distinct values fo $0,\gamma_0,\gamma_1$. \qed

\section{Iterative \name{} Variants}\label{sec:method:practice}
%\textbf{Grid Search for $\gamma$ vs. Bi-level Pptimization}
In the main body of the paper, we presented two \name{} variants, \name{}-M and \name{}-N. this section presents other practical extensions that are more flexible in their optimization approach but do not provide closed-form optimal solutions. 

Specifically, the \name{} variants discussed in the paper can be difficult to implement (especially \name{}-N) and are not flexible, e.g., considering other accuracy metrics or constraints requires a new theoretical study. Here, we discuss simple \name{} variants that use gradient descent to optimize MaxS-EFT (e.g., to solve the inner optimization in BiMax-M) and hyperparameter tuning to optimize MaxA-EFT (e.g., to solve the outer optimization in BiMax-M). This approaches can be less efficient and suboptimal, but can still provide accurate solutions, and are simple and flexible. In such \name{} variants, the learning rate and number of iterations act as knobs constraining how much the embeddings change. %We discuss two such \name{} variants.

\textbf{\name{}-IM}. First, note that MaxS-EFT is equivalent to minimizing the following loss
\begin{align}\label{eq:loss}
    \mathcal{L}=-\sum_{i\in[n_T]}\mQ_i\cdot (\mD_{Y_i}+\mDelta_{Y_i}).
\end{align} 
\name{}-IM performs gradient descent on loss in Eq.~\ref{eq:loss} with learning rate $\alpha$ for $t$ iterations ($\alpha$ and $t$ determined through hyperparameter tuning to maximize validation accuracy) with normalized gradients:
\begin{align*}%\label{eq:gd_rec}
    \mDelta^{(0)}\leftarrow \mathbf{0}, \;\mDelta^{(t)}\leftarrow\mDelta^{(t-1)}-\alpha\frac{\nabla_{\mDelta}\mathcal{L}}{\normx{\nabla_{\mDelta}\mathcal{L}}}.
 \end{align*}
\begin{lemma}\label{lemma:gd_nopet_equivalancy}
    \name{}-IM finds a solution equal to \name{}-IM, and thus optimally solves BiMax-M, whenever $\alpha t=\gamma^*$, where $\gamma^*$ is the optimal $\gamma$ value found by \name{}-M. 
\end{lemma}
\textit{Proof.} Observe that $\mG_i=-\nabla_{\mDelta_i}\mathcal{L}$, and therefore the gradient remains constant through optimization. Thus, after $t$ iterations, we have
$$
\mDelta_i^{(t)}=\alpha t \mathds{I}[\mG_i\neq0]\frac{\mG_i}{\normx{\mG_i}}.
$$
Thus, setting $\gamma^*=\alpha t$, we obtain the same results as \name{}-M.\qed

The above lemma implies using gradient descent with suitable $\alpha$ and $t$ can also provide accurate solutions, but at the cost of efficiency due to iterative updates and hyperparameter tuning  (instead of using the closed-form solutions), and the added challenge of finding suitable $\alpha$ and $t$. Appx.~\ref{sec:exp_iterative} presents an experimental study of these trade-offs.

\textbf{\name{}-IN}. Another alternative is \textit{\name{}-IN, an iteratively normalized \name{} variant}. \name{}-IN is similar to \name{}-IM but normalizes the embeddings after every update:
$$\mD^{(0)}_i\leftarrow \mD_i, \quad\mD^{(t)}_i\leftarrow \frac{\mD^{(t-1)}_i+\alpha \frac{\nabla_{\mDelta}\mathcal{L}}{\normx{\nabla_{\mDelta}\mathcal{L}}}}{\normx{\mD^{(t-1)}_i+\alpha \frac{\nabla_{\mDelta}\mathcal{L}}{\normx{\nabla_{\mDelta}\mathcal{L}}}}}\quad \forall i\in[n].$$

\name{}-IN ensure the fine-tuned embeddings are normalized similar to \name{}-N. Although, there is no theoretical equivalency between the solutions of \name{}-IN and \name{}-N, we observed similar accuracy in practice. Meanwhile, \name{}-IN is less efficient but simpler to implement.
\section{Multi-Label Formulation}\label{appx:multi_label}
In the main body of the paper, we provided fine-tuning solutions assuming each query has a single ground-truth answer. Here, we discuss how our results can be extended to a multi-label setting.  We first formalize the problem setting with multiple labels and then discuss how to extend our results.

\textbf{Ground-Truth Answers}. For a query, $\vq$, its ground-truth answer is a ranking of the data records so that the highest-ranked records are the most related to the query. %If $k$ data records are to be retrieved, the ground-truth answer is the top $k$ records based on the ranking.  
This ranking can be represented using \textit{relevance scores}, which, for each data record, quantifies how related the record is to the query (relevance score zero means the record is unrelated). The ground-truth ranking can be obtained by sorting the data records based on their relevance scores. More formally, we represent the ground-truth ranking for a query with a \textit{ground-truth rank index set}, $y=\{y_1,..., y_p\}$ and corresponding \textit{ground-truth relevance score set} $r=\{r_1, ..., r_p\}$ for some integer $p$ denoting the number of data records that are related to the query; the remaining records are not relate to the query. This means that record $\bar{D}_{y_{i}}$ has relevance $r_{i}$ to the query, and any record index not present in $y$ is assumed to have zero relevance score. Thus, sorting the dataset based on $r$ gives the ground-truth answer to the query. We drop the set $r$ if $|y|=1$, or if $r_1= ...=r_p=1$. 

\textbf{Fine-Tuning Query Sets}. For fine-tuning, a query set and corresponding ground-truth answers are available, consisting of queries $\bar{Q}$ and ground-truth answer sets $Y$, $R$, where for the $j$-th query $\bar{q}_j\in \bar{Q}$, $Y_j$ is the ground-truth index set and $R_j$ is the ground-truth relevance score set for $\bar{q}_j$. We assume this training set is split into two, a training set $\bar{Q}^T, Y^T$, $R^T$ and a validation set $\bar{Q}^V, Y^V, R^V$, with $n_T$ and $n_V$ queries respectively. Similar to single label setting, let $\mQ^T\in\mathcal{R}^{n_T\times d}$ and $\mQ^V\in\mathcal{R}^{n_V\times d}$ be matrices containing embeddings for training and validation queries. %For simplicity, we present most of our results assuming there is only one ground-truth answer or data record for any query, $q_j$, which we call the  \textit{correct answer} to $q_j$. We discuss extensions to multiple correct answers with relevance scores being available in Appx.~\ref{appx:multi_label}. The training set can be collected over time from user interactions with the system, by collecting labels, or by generating synthetic training data using LLMs (e.g., \cite{llamaindexDatasetGeneration, meng2022generating}). 

\textbf{Problem Formulation}. Both MaxS-EFT and MaxA-EFT can be modified to utilize multiple labels. For MaxS-EFT we can change the objective to $\sum_{i\in[n_T]}\sum_{j^*\in Y_i^T}\mQ^T_i\cdot (\mD_{j^*}+\mDelta_{j^*})$, where summation over $Y_i^T$ can optionally be weighted by relevance scores. For MaxA-EFT, we can adjust the inequalities in the definition of the correct answer to a query (i.e., Eq.~\ref{eq:I_iT}), so that for a query $\vq$ with two relevance scores $r_1$ and $r_2$, $r_1>r_2$, and for $R_1$ and $R_2$ containing document indexes with $r_1$ and $r_2$ relevance scores, we say $\vq$ is answered correctly when
\begin{align}\label{eq:correct_ans:multilabel2}
\vq\cdot(\mD_i+\mDelta_i)>\vq\cdot(\mD_j+\mDelta_j), \; i\in R_1, j\in R_2.
\end{align}

\textbf{\name{}}. To solve BiMax-M and BiMax-N, observe that the above modifications cause marginal changes for $\texttt{MaxS-M}$ and $\texttt{MaxS-N}$, and only require modifying the definition of $\mG$ in the corresponding optimal solutions. However, the solutions to the outer optimization problem in BiMax-M and BiMax-N require further modifications since now a different set of inequalities needs to be solved to find the range of $\gamma$ for which a query is answered correctly. However, each inequality is still of the same form as before (compare Eq.~\ref{eq:correct_ans:multilabel2} with Eq.~\ref{eq:I_iT}), and thus, the same methodology applies.

%MaxA-EFT and Constrained MaxA-EFT can also be similarly formulated to add a constraint per ground-truth answer and also enforce an ordering of the answers. For the sake of space, the exact statement of MaxA-EFT and Constrained MaxA-EFT with multiple labels, and the corresponding algorithm for solving them is deferred to Appendix~\ref{appx:multi_label}. 

\begin{table}[t]
\hspace{-2cm}
%\centering
        \begin{tabular}{c c c c c c c c}
        \toprule
       \textbf{Method} &\textbf{ArguAna}&\textbf{Fever}&\textbf{HotpotQA}&\textbf{NF-Corpus}&\textbf{NQ}&\textbf{SciFact}&\textbf{TriviaQA}
   \\\midrule
\name{}-M & \textbf{52.4  \textcolor{gray}{(+0)}}& \textbf{95.6  \textcolor{cadmiumgreen}{(+8.6)}}& \textbf{60.8  \textcolor{cadmiumgreen}{(+10.3)}}& 45.2  \textcolor{cadmiumgreen}{(+7.9)}& 50.6  \textcolor{cadmiumgreen}{(+9.5)}& 82.4  \textcolor{cadmiumgreen}{(+7.9)}& 40.1  \textcolor{cadmiumgreen}{(+9.9)}\\
\name{}-N & \textbf{52.4  \textcolor{gray}{(+0)}}& 93.1  \textcolor{cadmiumgreen}{(+6.1)}& 57.0  \textcolor{cadmiumgreen}{(+6.4)}& \textbf{46.9  \textcolor{cadmiumgreen}{(+9.6)}}& \textbf{56.2  \textcolor{cadmiumgreen}{(+15.1)}}& \textbf{88.3  \textcolor{cadmiumgreen}{(+13.8)}}& \textbf{45.0  \textcolor{cadmiumgreen}{(+14.7)}}\\
Adapter & \textbf{52.4  \textcolor{gray}{(+0)}}& 87.0  \textcolor{gray}{(+0)}& 53.0  \textcolor{cadmiumgreen}{(+2.5)}& 38.5  \textcolor{cadmiumgreen}{(+1.1)}& 41.1  \textcolor{gray}{(+0)}& \textbf{87.8  \textcolor{cadmiumgreen}{(+13.4)}}& 30.2  \textcolor{gray}{(+0)}\\
No Fine-Tuning & \textbf{52.4  }& 87.0  & 50.5  & 37.4  & 41.1  & 74.5  & 30.2  \\
\bottomrule
\end{tabular}
    \caption{NDCG@10 results for GTE-L on text datasets}
    \label{tab:res_gte_per_dataset}
\end{table}

\begin{table}[t]
\hspace{-2.2cm}
%\centering
        \begin{tabular}{c c c c c c c c}
        \toprule
       \textbf{Method} &\textbf{ArguAna}&\textbf{Fever}&\textbf{HotpotQA}&\textbf{NF-Corpus}&\textbf{NQ}&\textbf{SciFact}&\textbf{TriviaQA}
   \\\midrule
\name{}-M & \textbf{42.4  \textcolor{cadmiumgreen}{(+0.1)}}& \textbf{94.8  \textcolor{cadmiumgreen}{(+14.8)}}& \textbf{65.2  \textcolor{cadmiumgreen}{(+14.2)}}& \textbf{50.7  \textcolor{cadmiumgreen}{(+10.5)}}& 56.6  \textcolor{cadmiumgreen}{(+13.0)}& 86.5  \textcolor{cadmiumgreen}{(+11.1)}& 46.1  \textcolor{cadmiumgreen}{(+13.1)}\\
\name{}-N & \textbf{42.4  \textcolor{cadmiumgreen}{(+0.1)}}& 93.1  \textcolor{cadmiumgreen}{(+13.2)}& 63.4  \textcolor{cadmiumgreen}{(+12.3)}& 49.0  \textcolor{cadmiumgreen}{(+8.7)}& \textbf{59.6  \textcolor{cadmiumgreen}{(+16.0)}}& 89.8  \textcolor{cadmiumgreen}{(+14.4)}& \textbf{50.1  \textcolor{cadmiumgreen}{(+17.1)}}\\
Adapter & 41.7  \textcolor{burgundy}{(--0.6)}& 88.7  \textcolor{cadmiumgreen}{(+8.8)}& 54.5  \textcolor{cadmiumgreen}{(+3.4)}& 42.1  \textcolor{cadmiumgreen}{(+1.8)}& 42.4  \textcolor{burgundy}{(--1.2)}& \textbf{90.9  \textcolor{cadmiumgreen}{(+15.5)}}& 32.7  \textcolor{burgundy}{(--0.3)}\\
No Fine-Tuning & \textbf{42.3  }& 79.9  & 51.0  & 40.2  & 43.6  & 75.4  & 33.0  \\
\bottomrule
\end{tabular}
    \caption{NDCG@10 results for TE3-L on text datasets}
    \label{tab:res_openai_per_dataset}
\end{table}

\section{Additional Experiments and Details}\label{appx:exp}
Here we present additional experimental details and results:
\begin{itemize}
    \item Appx.~\ref{appx:exp:baselines_detail} discussed details on the implementation of Adaptor and PTFT, including hyper-parameter tuning, loss, and efficiency considerations.
    \item Appx.~\ref{appx:per_dataset} contains detailed per dataset results summarized in the paper's main body. 
    \item Appx.~\ref{sec:accuracy_during_training} presents experiments on the training processes of Adaptors and PTFT to understand their failure modes.
    \item Appx.~\ref{exp:norm:ablation} provides an ablation study of various normalization methods in \name{}.
    \item Appx.~\ref{sec:exp_iterative} provides an experimental comparison between \name{}-M and its corresponding iterative variant \name{}-IM.
\end{itemize}

\subsection{Adaptor and PTFT details and hyper-parameter tuning}\label{appx:exp:baselines_detail}
\textbf{-L loss.} The results presented with -L suffix use a modified version of MNR loss. The MNR loss, for a batch of $b$ queries, $\mQ^B$, and positive examples (i.e., data records that the correct answer to queries) $\mD^B$, where $\mD_i$ is a positive example for $\mQ_i$ for any $i\in[b]$, treats every other example in the batch as negative examples for $\mQ_i$. Let $\mathcal{S}_{i, j}=\frac{\mQ_i\cdot \mD_j}{\normx{\mQ_i}\normx{\mD_i}}$ for any $i, j\in[b]$. Then, MNR loss minimizes
$$
\mathcal{L}_{MNR}=-\sum_{i\in[b]}\log\frac{e^{\tau\mathcal{S}_{i, i}}}{\sum_{j\in[b]}e^{\tau\mathcal{S}_{i, j}}},
$$
for some temperature parameter $\tau$. Our modified loss, for the $i$-th query, ignores samples, $j$, for which $\mathcal{S}_{i, j}<\eta$ for some threshold $\eta$:
$$
\mathcal{L'}=-\sum_{i\in[b]}\log\frac{\mathds{I}[\mathcal{S}_{i, i}\geq\eta]e^{\tau\mathcal{S}_{i, i}}}{\sum_{j\in[b]}\mathds{I}[\mathcal{S}_{i, j}\geq\eta]e^{\tau\mathcal{S}_{i, j}}}.
$$
This follows the intuition that, at the fine-tuning stage, the model only needs to get better at distinguishing between records that have high similarity and is already accurate enough to separate relevant from non-relevant items. $\tau$ and $\eta$ are related and are jointly set through hyperparameter tuning.

\textbf{Modeling details and hyper-parameter Tuning.} For both Adaptor and PTFT, we did hyperparameter tuning to determine the learning rate (and use of a scheduler), batch size (although for PTFT it is bottlenecked by GPU memory size), number of training steps, model architecture and initialization (for Adaptor, we tried linear up to 8 layer MLPs) and which layers to train (for PTFT, we tried training the full model or training the last layer), and the choice and parameters of the loss function. We only performed hyper-parameter tuning for BGE-S, and used the resulting hyper-parameters for other models. We note that the choice of initialization is particularly important for Adaptor, and we observed a significant advantage to ensuring that at initialization, Adaptor is an identity function. For a single-layer adaptor, this can be achieved by setting the weight matrix to the identity matrix (also done by Llama Index~\citep{llamaindex_initialization}). For multi-layer adaptors, using ReLU activation, assuming queries are normalized (so that a query input $\vq\geq-1$), we achieve this by setting the weights to identity, the bias of first layer to \textbf{+1}, the bias of last layer to \textbf{-1} and other biases to \textbf{0}. We are unaware of this initialization being used by existing work for multi-layer adaptors, and we observed benefits to using this initialization over initialization that modify the embeddings at initialization (e.g., setting all biases to \textbf{0}).

\textbf{Efficient Implementation.} For Adaptor, our implementation is based on Llama Index~\cite{llamaindexFineTuningLlamaIndex} and for PTFT based on Sentence Transformers \cite{huggingfaceTrainingFinetuning}, but with additional considerations for initialization (for Adaptor, see above), loss function (see above) and efficiency. We observed that validation passes (for model checkpointing) often take longer than training passes (because the number of data records is often more than the number of training queries in our datasets), we used hyperparameter tuning to set the validation frequency to as low as possible without affecting final accuracy. For Adptor, when applying Adaptor only to queries (so that data records don't get embedded), we also used a vector index (Faiss library~\cite{douze2024faiss}) but did not observe any speed-ups (perhaps due to the already parallelizable nature of answering batched queries, and that we only do top-1 lookup during validation). For PTFT, a single validation pass, which requires re-embedding the entire dataset, can take more than an hour on our large datasets (Fever, NQ, TriviaQA, HotpotQA). To reduce the computational cost, for each validation query, we selected its top-10 answers based on the pre-trained model and only included those (in addition to ground-truth training and validation answers) data records in the dataset for validation. This provided more than 10x speed-up for validation on large datasets, and we observed similar final accuracy (intuitively, this removes data records from validation that, unless the pre-trained model significantly changes, are unlikely to impact validation accuracy).  Finally, we use early stopping if validation accuracy drops by more than 5\% compared with the maximum it had achieved. 

%Tables~\ref{tab:res_gte_per_dataset}-~\ref{tab:res_clibl_per_dataset} show per dataset results using GTE-L, TE3-L, CLIP-B and CLIP-L models for text and image datasets. 

\subsection{Other Per Dataset Results}\label{appx:per_dataset}
Tables~\ref{tab:res_gte_per_dataset}-\ref{tab:res_clibl_per_dataset} present the per dataset results for the embedding models GTE-L, TE3-L, CLIP-B and CLIP-L. The tables (in addition to Table~\ref{tab:res_bge_per_dataset}) present the detailed results from which Tables~\ref{tab:res_text_avg}-\ref{tab:res_img_avg} are generated.

\begin{table}[t]\centering
\begin{minipage}{0.49\textwidth}
        \begin{tabular}{c c c }
        \toprule
       \textbf{Method} &\textbf{COCO}&\textbf{Flickr}
   \\\midrule
\name{}-M & \textbf{29.7  \textcolor{cadmiumgreen}{(+11.6)}}& \textbf{52.2  \textcolor{cadmiumgreen}{(+16.5)}}\\
\name{}-N & \textbf{29.9  \textcolor{cadmiumgreen}{(+11.9)}}& \textbf{52.4  \textcolor{cadmiumgreen}{(+16.7)}}\\
Adapter & 20.0  \textcolor{cadmiumgreen}{(+2.0)}& 39.8  \textcolor{cadmiumgreen}{(+4.1)}\\
No Fine-Tuning & 18.0  & 35.7  \\
\bottomrule
\end{tabular}
    \caption{NDCG@10 results for CLIP-B on image datasets}
    \label{tab:res_clibb_per_dataset}
    \end{minipage}
    \hfill
\begin{minipage}{0.49\textwidth}
        \begin{tabular}{c c c }
        \toprule
       \textbf{Method} &\textbf{COCO}&\textbf{Flickr}
   \\\midrule
\name{}-M & \textbf{31.4  \textcolor{cadmiumgreen}{(+9.0)}}& \textbf{55.0  \textcolor{cadmiumgreen}{(+12.5)}}\\
\name{}-N & \textbf{31.5  \textcolor{cadmiumgreen}{(+9.1)}}& \textbf{55.1  \textcolor{cadmiumgreen}{(+12.6)}}\\
Adapter & 25.0  \textcolor{cadmiumgreen}{(+2.6)}& 48.0  \textcolor{cadmiumgreen}{(+5.5)}\\
No Fine-Tuning & 22.4  & 42.5  \\
\bottomrule
\end{tabular}
    \caption{NDCG@10 results for CLIP-L on image datasets}
    \label{tab:res_clibl_per_dataset}
    \end{minipage}
\end{table}

\begin{figure}[t]
    \centering
    \includegraphics[width=1\linewidth]{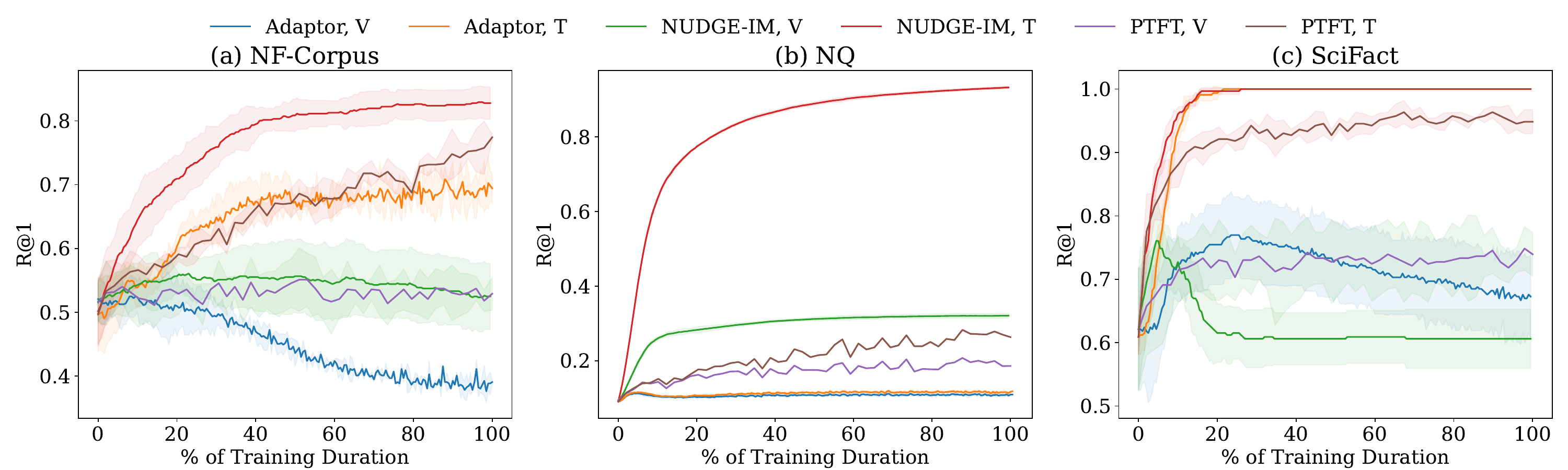}
    \caption{Training and validation accuracy for BGE-S on three datasets}
    \label{fig:bges_over_training}
\end{figure}

\subsection{Training and Validation Accuracy During Training}\label{sec:accuracy_during_training}

To better understand the differences and failure modes of the approaches, Fig.~\ref{fig:bges_over_training} shows the validation and training accuracy for Adaptor, PTFT and \name{}-IM, where T and V respectively signify training and validation sets. 

Fig.~\ref{fig:bges_over_training} (a) and (b) show two failure modes for Adaptor, overfitting, and underfitting. Specifically, Fig.~\ref{fig:bges_over_training} (a) shows \name{} provides much better validation accuracy compared with Adaptors at the same training accuracy, suggesting that  Adaptor simply overfits to the training set instead of learning generalizable patterns. Fig.~\ref{fig:bges_over_training} (b) shows the other end of the spectrum, where Adaptor fails to fit the training set at all (while \name{} both fits the training set and improves validation accuracy). We observed this behavior on large data and query sets. We also observed (but not shown here) that increasing the number of parameters, e.g., by introducing additional layers, did not improve accuracy, suggesting that perhaps using adaptors is a wrong modeling choice.  Fig.~\ref{fig:bges_over_training} (c) shows the only dataset where Adaptor performs well, where it both fits the training set and improves validation accuracy. 

PTFT, on the other hand, has a much smaller generalization gap compared with Adaptor. However, both validation and training accuracy increase at a much slower pace, and eventually plateau. Especially for NQ, we observe that the model underfits the training set. We hypothesized that one reason could be due to the loss function used, where indeed Table.~\ref{tab:res_bge_per_dataset} shows our attempt at modifying the loss function does help improve accuracy on NQ, but not consistently across datasets (and worsens the accuracy on other datasets). 

%Nonetheless, the observed rate of convergence is slow, and the model fails to make meaningful progress after hours of training. Although we do not rule out the possibility that spending more computational resources and/or improving the loss function can improve the performance on NQ, the current results show our best effort to present results based on the state-of-the-art training procedure within our available resources. 

\subsection{Normalization Ablation Study}\label{exp:norm:ablation}
We compare \name{}-N, with \name{}-IN (as described in Sec.~\ref{sec:method:practice}) with two other potential variants to understand the impact of normalization. \name{}-M+N is a variant that first performs \name{}-M and then normalizes embeddings post-hoc. \name{}-IM+R is a variation of \name{}-IM with L2 regularization added to the loss to penalize embeddings with large norms. Table~\ref{tab:norm_ablation}  shows the results for this experiment, showing that normalizing embeddings after optimization, i.e., \name{}-M+N performs worse than when normalization is considered as part of optimization, which is the case for both \name{}-IN and \name{}-N. \name{}-IM+R performs worse than all methods, showing an advantage for enforcing a normalization constraint over L2 regularization. Meanwhile, \name{}-IN and \name{}-N perform similarly (see Sec.~\ref{sec:method:practice} for a discussion between the two).

\begin{table}[t]
%\begin{minipage}{0.3\textwidth}
\centering
        \begin{tabular}{c c}
        \toprule
      \textbf{Method} & \textbf{NDCG@10}
   \\\midrule
\name{}-N & \textbf{61.1}\\
\name{}-IN & \textbf{61.8} \\
\name{}-M+N & 58.8\\
\name{}-IM+R & 52.4
\\\bottomrule
\end{tabular}
    \caption{Normalization Study, avg. BGE-S accuracy}
    \label{tab:norm_ablation}
%\end{minipage}
\end{table}

\begin{table}[t]
%\hspace{-1.5cm}
\centering
        \begin{tabular}{c c c c c }
        \toprule
       \textbf{Method} &\textbf{R@1}&\textbf{R@10}&\textbf{NDCG@10}&\textbf{Fine-Tuning Time (s)}
   \\\midrule
\name{}-IM-$(10^{-4}, 10^3)$ & \textbf{50.4}& \textbf{66.5}& \textbf{57.2} & 11.3\\
\name{}-IM-$(10^{-3}, 10^2)$ & \textbf{50.4}& \textbf{66.5}& \textbf{57.1} & 4.16\\
\name{}-IM-$(10^{-2}, 10)$ & 48.0& 59.2& 51.1 & \textbf{3.46}\\
\name{}-M & \textbf{50.5}& \textbf{66.6}& \textbf{57.3} & \textbf{3.49}\\
\bottomrule
\end{tabular}
    \caption{Accuracy/efficiency of iterative and non-iterative \name{} variants (BGE-S on text datasets)}
    \label{tab:res_opt}
\end{table}

\subsection{Iterative vs. Closed-Form \name{} variants}\label{sec:exp_iterative}
We present results comparing \name{}-M and \name{}-IM. We use \name{}-IM-($\alpha$, $t$) to refer to \name{}-IM with learning rate $\alpha$ run for $t$ iteration (although we use model checkpointing based on validation accuracy, so the presented accuracy can be for a model trained with fewer iterations than $t$). Overall, the results show that \name{}-M saves time while providing the best accuracy, while some \name{}-IM variants are as accurate while being less efficient. Nonetheless, \name{}-IM variants are simple to implement and more flexible (e.g., in terms of validation accuracy metric to use), and thus may be preferred in some applications.

\end{document}